# Fractal and Chaotic Activation Functions in Echo State Networks: Preprocessing Topology Governs the Echo State Property

Rae Chipera[a, b, *], Jenny Du[a], Irene Tsapara[a]

[a] *National University School of Technology and Engineering, 9388 Lightwave Ave, San Diego California, 92123 USA*

[b] *Jaxorik AI Research Group, 6300 Riverside Plaza Ln. NW #100, Albuquerque, New Mexico, 87120, USA*

[*] Corresponding author.

E-mail addresses: rachipe@jaxorik.com, zdu@nu.edu, itsapara@nu.edu.

URLs: https://nu.edu, https://jaxorik.com

**Abstract**

Contemporary reservoir computing relies heavily on globally Lipschitz, well-behaved activation functions, limiting applications in defense, disaster response, and pharmaceutical modeling where robust operation under extreme conditions is critical. We systematically investigate non-smooth activation functions, including chaotic, stochastic, and fractal variants, in echo state networks. Through comprehensive parameter sweeps across 36,610 reservoir configurations, we demonstrate that several non-smooth functions not only maintain behavior consistent with the Echo State Property (ESP) but outperform traditional smooth activations in convergence speed and spectral radius tolerance.

Notably, the Cantor function (continuous everywhere and zero derivative almost everywhere) maintains ESP-consistent behavior up to spectral radii of $\rho \approx 10$, an order of magnitude beyond typical bounds for traditional functions, while achieving 2.6x faster convergence than tanh and ReLU. We introduce a theoretical framework for quantized activation functions, defining a Degenerate Echo State Property (d-ESP) that captures stability for discrete-output functions and proving that d-ESP implies traditional ESP.

We conjecture a critical crowding ratio $Q = \frac{N}{k}$ (reservoir size / quantization levels) that predicts failure thresholds for discrete activations. Our analysis reveals that preprocessing topology, rather than continuity per se, determines stability: monotone, compressive preprocessing maintains ESP across scales, while dispersive or discontinuous preprocessing triggers sharp failures. While our findings challenge assumptions about activation function design in reservoir computing, the mechanism underlying the exceptional performance of certain fractal functions is only partially explained by the effective-gain analysis presented here, suggesting fundamental gaps in our understanding of how geometric properties of activation functions influence reservoir dynamics.

**Keywords:**
Reservoir computing, Echo State Networks, Fractal Activation Functions, Echo State Property, Chaotic dynamics, machine learning

## 1. Introduction

Contemporary recurrent networks, including reservoir computing (RC), almost exclusively employ smooth, globally Lipschitz continuous activation functions (e.g. **Jaeger, 2001; Maass et al., 2002; Lukoševičius and Jaeger, 2009; Lukoševičius, 2012**). This restriction to well-behaved functions is largely motivated by stability concerns and analytical tractability. However, complex systems theory suggests that unconventionally-constructed functions, which we define as (i) chaotic, (ii) stochastic, and/or (iii) fractal, may provide computational advantages. We ask: what happens when we deliberately introduce non-smooth activation functions into RC?

The relationship between chaos and computation has long fascinated researchers. Following the analysis of chaotic dynamics in random neural networks by **Sompolinsky et al. (1988), Bertschinger and Natschläger (2004)** demonstrated that randomly connected networks of threshold gates perform complex real-time computation on time series only near the critical boundary between ordered and chaotic dynamics. While that work concerns the network's dynamical regime, it motivates our question of whether irregularity in the activation function itself can be computationally advantageous. **Boedecker et al. (2012)** extended this analysis to echo state networks (ESNs) examining information processing capacity at critical parameter values. However, **Carroll (2020)** simulated a range of reservoir computers and found that optimal performance does not reliably coincide with the edge of chaos, with some parameter regimes at criticality performing poorly. Reservoir computers have meanwhile been applied to signal classification and low-probability-of-detection communications **(Carroll, 2017, 2021)**, domains in which robust operation under constrained parameter ranges is a practical requirement rather than a theoretical convenience. This highlights a gap between theoretical predictions and practical requirements where systems must balance computational efficiency with robust performance under extreme conditions.

These conflicting findings may stem from the exclusive use of globally Lipschitz, well-behaved activation functions in prior work, highlighting the need for systematic empirical investigation of irregular dynamics in ESNs. We hypothesize that non-smooth activation functions, with their fundamentally different dynamics, may behave differently at this critical transition.

Our results challenge conventional wisdom about activation function design. We find that fractal activations can significantly accelerate convergence in ESNs while maintaining stability. Most surprisingly, the Cantor function maintains ESP-compliant behavior up to spectral radii $\rho \approx 10$, up to an order of magnitude beyond the typical stable range of smooth activations. That same function maintains partial convergence up to $\rho = 100$. Eigenvalue verification of this is provided in **Section 6**. These findings suggest that non-smooth dynamics may actually be advantageous for RC.

Our contributions are fourfold:

1. Formulation: We formalize classes of irregular activations and specify real-valued constructions suitable for RC.
2. Diagnostics: We adopt ESP-relevant convergence metrics and finite-time decay-rate estimates to assess contraction under inputs with non-zero practical probability mass (Gaussian, uniform, and sparse specifically chosen to avoid the zero-probability-mass pathologies identified by **Yildiz et al., 2012**).
3. Phase Diagrams: We empirically map ESP compliance across reservoir size $N$, spectral radius $\rho$, and leak rate $a$, using large multi-seed sweeps.
4. Theory: We introduce the degenerate Echo State Property (d-ESP) for quantized activations, prove d-ESP implies ESP under the leaky update, and conjecture a crowding ratio $Q = \frac{N}{k}$ that predicts quantization-induced failure.

Taken together, our results reveal scale-dependent regimes: some non-smooth activations maintain ESP-consistent behavior and, in certain settings, exhibit faster state convergence than standard smooth activations, while others break ESP as $N$ grows. We discuss implications for activation design in RC.

## 2. Theoretical Framework

Beyond the edge of chaos debate, ESNs have proven remarkably robust to various forms of noise and irregularity (**Wikner et al., 2024**). The Echo State Property (ESP) is fundamental to their success.

### *2.1 Echo State Property: Classical Theory and Extensions*

Intuitively, ESP ensures that the reservoir acts as a fading memory system: the network's current state depends only on its recent input history, not on how it was initialized or its distant past. This ensures that the model does not get stuck in internal feedback loops or make its behavior unpredictable (**Jaeger, 2001**).

However, recent work reveals that ESP is more complex than originally thought. Traditionally and as defined by **Jaeger (2001)**, a reservoir is said to satisfy the echo state property (ESP) if: $||x(t) - x'(t)|| \to 0$ as $t \to \infty$ for any two state trajectories with the same input but different initial conditions, where $|| \cdot ||$ denotes the Euclidean norm. In practice, spectral radius $\rho < 1$ has been treated as a heuristic sufficient condition for ESP under smooth, globally Lipschitz activations **(Jaeger, 2001)**, though subsequent work shows this condition is neither necessary nor sufficient in general.

**Yildiz et al. (2012)** and **Manjunath and Jaeger (2013)** demonstrated that ESP depends not only on spectral radius but also on the input distribution and activation function properties. **Yildiz et al. (2012)** proved that pathological input distributions with zero probability mass can violate ESP even when $\rho < 1$, while **Manjunath and Jaeger (2013)** showed that ESP can be maintained for certain input classes even when $\rho \geq 1$, provided the activation function and input scaling satisfy specific contractivity conditions. Meanwhile, **Grigoryeva & Ortega (2018)** established that ESP is a prerequisite for universal approximation in RC, highlighting its importance beyond just stability.

These findings suggest that smoothness may not be strictly necessary for ESP compliance. Moreover, the classical sufficient conditions are known to be conservative: **Buehner and Young (2006)** sharpened the standard singular-value bound via a weighted operator norm, while **Verstraeten et al. (2007)** and **Ozturk et al. (2007)** demonstrated empirically that reservoir stability depends on the interplay between activation function properties, input scaling, and network topology. **Appeltant et al. (2011)** further showed that even highly nonlinear physical reservoirs with delay dynamics can exhibit ESP-consistent stability. Together, these results indicate a substantial gap between what is provably sufficient for ESP (global Lipschitz bounds on the update map) and what is actually necessary, motivating our question: can bounded activations far outside this classical regime maintain ESP through alternative mechanisms?

Since RC architectures train only the linear readout layer, we avoid the computational challenges of backpropagation through time while maintaining the ability to capture complex nonlinear dynamics. Sufficient conditions for ESP tie the reservoir update's global Lipschitz constant to architectural hyperparameters (spectral radius, leak rate, and input scaling.) This work asks whether non-smooth activations can be used in ESNs without violating ESP, and whether they offer practical benefits. We organize unconventionally constructed activations into three families: (i) **chaotic** (logistic map with sensitive dependence on initial conditions), (ii) **stochastic** (Brownian motion-driven with inherent randomness), and (iii) **fractal** (Weierstrass, Cantor, and Mandelbrot variants with self-similar structure).

*2.2 Echo State Property Versus Boundedness*

Our hypothesis explores whether bounded non-smooth functions may maintain ESP not through global Lipschitz continuity, but through alternative contraction mechanisms. We investigate which specific properties beyond boundedness determine stability.

**Proposition 2.3 Global Boundedness and Absorbing Set.**

Traditional reservoir computing relies on spectral radius $\rho < 1$ to prevent state explosion. However, bounded activation functions guarantee that reservoir states remain confined regardless of how large $\rho$ becomes. If the activation $f$ maps to the interval $[L, U]$, then the reservoir states cannot diverge, even with arbitrarily large spectral radii.

**Intuition:** The leak term $(1-a), a \in (0,1]$ causes old states to decay geometrically, while the activation term $af(\cdot)$ can only add bounded increments. This creates a "bounded basin" that traps all trajectories. Let $B = \max\{|L|, |U|\}$ and consider the leaky update:

$$x_t = (1-a)\, x_{t-1} + af\left(W^{in} u_t + W^{res}\, x_{t-1}\right) \tag{1}$$

where $f$ applies component-wise: $f(x)_i = f(x_i)$ .

Then, for any $x_0 \in \mathbb{R}^N$ and any input sequence, $\limsup_{t\to\infty} \left|\left|x_t\right|\right|_\infty \le B$; moreover, if $\left|\left|x_0\right|\right|_\infty \le B$, then $\left|\left|x_t\right|\right|_\infty \le B$ for all $t \ge 1$ regardless of $\rho(W^{res})$.

**Proof:** Let $B = \max\{|L|, |U|\}$. Elementwise boundedness gives $\left|\left|f(\cdot)\right|\right|_\infty \le B$. The leaky update yields:

$$\left|\left|x_t\right|\right|_\infty \le (1-a)\left|\left|x_{t-1}\right|\right|_\infty + aB.$$

This linear recurrence has the closed form:

$$\left|\left|x_t\right|\right|_\infty \le (1-a)^t \left|\left|x_0\right|\right|_\infty + B(1-(1-a)^t)$$

Since $(1-a)^t \to 0$ as $t \to \infty$, the right-hand side converges to $B$, hence $\limsup_{t\to\infty} \left|\left|x_t\right|\right|_\infty \le B$. Moreover, if $\left|\left|x_0\right|\right|_\infty \le B$, the same closed form gives $\left|\left|x_t\right|\right|_\infty \le (1-a)^t B + B[1-(1-a)^t] = B$ for all $t$. The hypercube $\mathcal{H} = [-B, B]^N$ is absorbing regardless of $\rho(W^{res})$. ■

**Note:** While we use $L_\infty$ here for analytical tractability, our empirical tests (**Section 5**) employ $L_2$, which is standard in reservoir computing and measures aggregate system convergence. In finite dimensions, all norms are equivalent up to constants, so boundedness in $L_\infty$ implies boundedness in $L_2$.

**Remark. Boundedness vs. ESP.** This guarantees states stay finite but does not ensure convergence to unique trajectories (ESP), even for large $\rho$. Boundedness prevents divergence; ESP requires additional contractivity conditions. This distinction is crucial: boundedness alone ensures global state confinement but does not eliminate dependence on initial conditions. Thus, boundedness provides necessary but not sufficient structure for ESP.

*2.4 Degenerate Echo State Property and Critical Quantization*

We define a Degenerate Echo State Property (d-ESP) and show that d-ESP implies traditional ESP.

**Definition 2.4.1 Quantized Activation Function.**

An activation function $f: \mathbb{R} \to L$ is k-quantized if $|L| = k \geq 2$ where $L \subset \mathbb{R}$ is the finite set of possible output values (the codebook). Extend $f$ elementwise to $f: \mathbb{R}^N \to L^N$. The codebook geometry is characterized by two quantities: $D_L \coloneqq \max_{\ell,\ell' \in L} |\ell - \ell'|$, $\delta_L \coloneqq \min_{\ell \neq \ell' \in L} |\ell - \ell'|$. Since $L$ is finite with $k \geq 2$ elements, both quantities are finite and $\delta_L > 0$. Intuitively, $D_L$ is the diameter of the codebook (the largest possible jump between activation levels), while $\delta_L$ is the resolution (the smallest gap separating distinct levels).

**Definition 2.4.2 Degenerate Echo State Property.**

Traditional ESP is formulated as asymptotic convergence of state trajectories. For quantized activations, the natural object of study is instead the symbol sequence because outputs take values in a discrete codebook. The activation outputs of two trajectories cannot approach each other gradually; at each step they either coincide or differ by at least $\delta_L$. This motivates a convergence notion stated directly in terms of eventual symbol agreement.

We therefore propose a Degenerate ESP (d-ESP) that asks: do the activation outputs eventually produce the same discrete patterns? If two reservoirs start from different initial conditions but eventually generate identical sequences of activation levels, they can be considered converged for practical purposes. The reservoir still has a consistent memory, but it expresses that memory through discrete symbols rather than continuous values.

For the leaky ESN from **Eq. 1** let $\mathcal{A}_t = W^{in}u_t + W^{res}x_{t-1}, s_t = f(\mathcal{A}_t)$ and likewise $s'_t$ for a second trajectory. The system has d-ESP if for any $x_0, x'_0$ driven by the same input sequence $\{u_t\}$ there exists $T < \infty$ such that $s_t = s'_t$ for all $t \geq T$. Equivalently, the two trajectories generate identical symbol sequences after finite time.

**Definition 2.4.3 Crowding Ratio.**

For a k-quantized activation function in a reservoir of size N, we define the crowding ratio as:

$$Q = \frac{N}{k} \tag{2}$$

where $k$ is the number of discrete levels.

**Conjecture 2.4.4 Critical Crowding Ratio.**

We conjecture that for a fixed codebook geometry $(D_L, \delta_L)$, leak rate, and spectral radius, there exists $Q^*$ such that a k-quantized activation satisfies d-ESP for $Q < Q^*$ and begins failing for $Q > Q^*$. Further, we conjecture that $Q^*$ varies with these parameters.

**Lemma 2.5 d-ESP is Equivalent to the Natural Norm Convergence Statement.**

Let $f$ be k-quantized with codebook $L \subset \mathbb{R}$ (i.e., $L$ is the set of possible activation values) and minimal separation $\delta_L > 0$. For any two trajectories driven by the same input sequence with symbol sequences $s_t = f(\mathcal{A}_t)$ and $s_{t'} = f(\mathcal{A}_{t'})$,

$$||s_t - s_t'|| \to 0 \iff \text{There exists } T < \infty \text{ such that } s_t = s_t' \text{ for all } t \geq T \tag{3}$$

**Proof:** ($\Leftarrow$) is immediate. ($\Rightarrow$): wherever $s_t \neq s_t'$, they differ in at least one coordinate by at least $\delta_L$, so $||s_t - s_t'|| \geq \delta_L$ in any p-norm. Convergence to zero therefore forces $s_t = s_t'$ for all sufficiently large $t$. ■

**Theorem 2.6 d-ESP Implies ESP.**

Suppose the leaky ESN **(Eq. 1)** with k-quantized activation $f$ satisfies d-ESP. For any initial conditions $x_0, x_{0'}$ and any input sequence, there exists $T < \infty$ with $s_t = s_t'$ for all $t \geq T$. Then $||x_t - x_t'|| \to 0$ for every such pair of trajectories.

That is, the system satisfies ESP.

**Proof:** For $t > T$, both trajectories receive identical activation outputs, so the difference $\Delta_t = x_t - x_t'$ evolves as $\Delta_t = (1-a)\Delta_{t-1}$, giving $||\Delta_t|| = (1-a)^{t-T}||\Delta_T||$. For $a \in (0,1)$, this converges to zero geometrically; for $a = 1$, $\Delta_t = 0$ for all $t > T$. ■

**Lemma 2.7 Collision-Driven Contraction Bound.**

We now establish a contraction bound that governs how state differences evolve in quantized systems. When quantized outputs match, we observe exponential contraction. When they differ, contraction is limited by the maximum distance between quantization levels.

For two trajectories with difference $\Delta_t = x_t - x_t'$, and quantized outputs $s_t = f(\mathcal{A}_t), s_t' = f(\mathcal{A}_t')$, the state difference satisfies:

$$||\Delta_t||_\infty \leq (1-a)||\Delta_{t-1}||_\infty + aD_L\mathbf{1}\{s_t \neq s_t'\} \tag{4}$$

where $\mathbf{1}\{\cdot\}$ is the indicator function (1 if the condition is true, 0 otherwise).

**Remark. Iterated bound.** Unrolling **Eq. 4** from an arbitrary $\Delta_0$ gives $||\Delta_t||_\infty \leq (1-a)^t||\Delta_0||_\infty + aD_L \sum_{j=1}^{t}(1-a)^{t-j}m_j$ where $m_j \coloneqq \mathbf{1}\{s_j \neq s_j'\}$. The first term decays geometrically and is independent of the codebook. The second is a geometrically discounted count of symbol mismatches. Convergence is therefore governed not by whether mismatches occur, but by how persistently they recur. Since $\sum_{j=1}^{t}(1-a)^{t-j} \leq \frac{1}{a}$, a sequence of mismatches with long-run discounted $p$ yields $limsup_{t\to\infty}||\Delta_t||_\infty \leq pD_L$.

The steady-state error is thus bounded by the product of mismatch density and codebook diameter. d-ESP is the case $p = 0$, for which the bound recovers exponential convergence to zero; a persistent $p > 0$ produces a nonvanishing error floor, which is the behavior observed for the discrete function tested here, as shown in **Section 6.2.2.**

**Remark. Norm Conversion.** Convergence in Section 5 is measured in $L_2$, which is standard in reservoir computing. Since $||\Delta_t||_2 \leq \sqrt{N}||\Delta_t||_\infty$, the bound above converts directly, at the cost of a dimensional factor $\sqrt{N}$. This factor is not innocuous when comparing across reservoir sizes: part of any increase in $L_2$ steady-state error with $N$ reflects the norm conversion rather than degraded contraction, and the two effects should be separated before attributing growth to quantization alone.

**Proof:** When $s_t = s_t'$, we have $f(\mathcal{A}_t) = f(\mathcal{A}_t')$, so:

$||\Delta_t|| = \left||(1-a)\Delta_{t-1} + a\big(f(\mathcal{A}_t) - f(\mathcal{A}_t')\big)\right|| = (1-a)||\Delta_{t-1}||.$

When $s_t \neq s'_t$, the activation outputs differ. Since $f$ maps to finite codebook $L$, we define the worst-case separation between distinct quantization levels: $D_L = \max_{\ell,\ell' \in L} ||\ell - \ell'||$, which gives us: $||f(\mathcal{A}_t) - f(\mathcal{A}'_t)|| \leq D_L$.

Therefore: $||\Delta_t|| \leq (1-a)||\Delta_{t-1}|| + a||f(\mathcal{A}_t) - f(\mathcal{A}'_t)|| \leq (1-a)||\Delta_{t-1}|| + aD_L$. ∎

*2.8 ESP Mechanisms for Continuous Non-Smooth Functions*

The d-ESP framework (**Section 2.4**) applies to quantized activations. For continuous but non-smooth functions, alternative contraction mechanisms may ensure ESP. The following are some candidates:

1. **Hölder continuity:** Functions satisfying $|f(x) - f(y)| \leq C|x - y|^H$ for $H \in (0,1)$ are not Lipschitz but may still contract under composition with leaky integration.
2. **Monotonicity:** Bounded monotonic functions preserve ordering and may induce contraction through order-theoretic properties.
3. **Average-case contraction:** Even nowhere-differentiable functions may contract in expectation over realistic input distributions.

Note that mechanisms 2 and 3 are examined empirically in Sections 7.3 and 7.4 respectively, while rigorous characterization of Hölder condition remains future work. All bounded functions tested satisfy **Proposition 2.3**, preventing divergence even at large $\rho$.

*2.9 Fading Memory Property in Non-Smooth Activations*

The fading memory property is a fundamental characteristic of reservoir computing systems that ensures the network's current state depends primarily on recent inputs, with influence from distant past inputs decaying exponentially over time. **(Jaeger, 2001; Gonon & Ortega, 2021).** While ESP guarantees that two trajectories with identical inputs converge regardless of initial conditions, the fading memory property specifically characterizes how temporal dependencies decay.

For a leaky ESN with the update **(Eq. 1)**, the fading memory property requires that the influence of input $u(s)$ on state $x(t)$ decays as $|t - s|$ increases. Formally, for any two input sequences $u$ and $u'$ that differ only at time $s$, the state difference $||x_u(t) - x_{u'}(t)||$ should decay exponentially with $(t - s)$.

**Gonon & Ortega (2021)** established that fading memory is equivalent to ESP for ESNs with sufficiently regular activation functions. However, the relationship between fading memory and ESP for non-smooth activations has not been thoroughly characterized. We will evaluate this empirically.

*2.10 Input-to-State Stability in ESNs*

Input-to-State Stability (ISS) is a fundamental concept in dynamical systems theory that characterizes how bounded inputs lead to bounded state trajectories with bounded sensitivity to perturbations. For RC, ISS ensures that finite-amplitude input signals cannot drive the network into unbounded or chaotic regimes, regardless of spectral radius or initial conditions.

A dynamical system is ISS if there exist functions $\beta$ (class-KL) and $\tau$ (class-K∞) such that for any initial condition $x(0)$ and any bounded input sequence u:

$$||x(t)|| \leq \beta\left(||x(0)||, t\right) + \tau(sup_{s \leq t}||u(s)||) \tag{5}$$

Where $\beta(r, t) \to 0$ as $t \to \infty$ for fixed $r$ (decay of initial conditions) and $\tau$ is a monotonically increasing function (bounded inputs → bounded states).

For the leaky ESN with update **(Eq. 1)** ISS requires three criteria. 1. Initial condition influence decays to zero (ensured by the leak rate term $a > 0$), 2. Bounded inputs produce bounded states, and 3. State sensitivity to input perturbations is bounded.

**Theorem 2.10.1 Practical ISS for Bounded Monotonic Activations.**

Let $\gamma: \mathbb{R} \to [-B, B]$ be a bounded monotonic activation function and consider the leaky ESN with leak rate $a \in (0,1]$. Then the system satisfies practical (bias) ISS with respect to bounded input sequences. There exists a class-$\mathcal{KL}$ function $\beta$, a class-$\mathcal{K}_\infty$ function $\tau$, and a constant bias $c \geq 0$ such that $||x(t)|| \leq \beta\left(||x(0)||, t\right) + \tau\left(\sup_{s \leq t}||u(s)||\right) + c$ with $\beta(r, t) = (1 - a)^t r, \tau(u) = \frac{B||W_{in}||}{a} u$, and bias $c = \frac{B||W||}{a}$.

**Proof:** Given bounded activation $\gamma(\cdot) \in [-B, B]$, we have from **Proposition 2.3:**

$||x(t) \leq (1-a)^t||x(0)|| + aB\left(||W|| + ||W_{in}||\ U_{max}\right) \sum_{\{k=0\}}^{\{t-1\}}(1-a)^k$ where $U_{max} = sup_s||u(s)||$ is the maximum input norm. The geometric series sums to $\frac{1-(1-a)^t}{a} \leq \frac{1}{a}$. Therefore $||x(t)|| \leq (1-a)^t\ ||x(0)|| + B\ (||W|| + ||W_{in}||\ U_{max})$. This gives us the ISS bound with: $\beta(r,t) = (1-a)^t\ r$ (exponential decay of initial conditions) and $\tau(u) = B\ \left(||W|| + ||W_{in}||\ u\right)$ (linear growth of input bound.) ■

The key observation is that boundedness of the activation function is sufficient for ISS, independent of Lipschitz continuity or differentiability.

**Remark.** The bias term $c$ arises from the reservoir's own recurrent weight norm and is independent of the input; it does not vanish as $t \to \infty$, so this practical ISS (bounded-bias ISS) rather than the zero-bias ISS definition requiring $\tau(0) = 0$. This is standard in reservoir computing, where the recurrent term itself contributes a persistent, but bounded offset.

## 3. Non-Smooth Activation Functions Defined

We define non-smooth activation functions into one of the categories: fractal, chaotic, or stochastic. We include functions to explore a broad range of non-smooth dynamics including:

- Continuity vs. discontinuity (Cantor function vs. Cantor set, two different versions of the Mandelbrot function, and two different versions of the logistic map),
- Monotonic vs. non-monotonic (Cantor function vs. Weierstrass)
- Deterministic vs. stochastic (logistic map, Mandelbrot, Cantor, and Weierstrass are all deterministic; Brownian motion introduces genuine stochasticity)
- Fractal vs. chaotic vs. stochastic: Mandelbrot, Weierstrass and Cantor embody fractal geometry; logistic map represents chaotic iteration; Brownian motion introduces stochastic perturbation.

This will garner an appropriate understanding of the entire space of non-smooth functions as activations in neural networks.

**Remark.** For vector inputs $x \in \mathbb{R}^n$, these functions are applied elementwise, except for Brownian motion which generates a vector of independent increments.

**Note:** We predict that Brownian motion's stochastic nature will fail to satisfy ESP by any definition (in other words, we hypothesize that it will never converge). However, we are including it as a negative control.

*3.1 Mandelbrot-Based Activation Functions*

The Mandelbrot set, as defined by **Mandelbrot (1980),** is an iterative complex equation:

$$z_{n+1} = z_n^2 + c$$

where $z$ is a complex number initialized at 0 ($z_0 = 0$), and $c$ is a point in the complex plane. The Mandelbrot set consists of all values of $c$ for which the iteration remains bounded (does not diverge to infinity). The Mandelbrot activation function uses escape time dynamics. For the complex parameter $c$, the escape time $T(c)$ counts how many iterations it takes for $|z_n|$ to exceed the escape criterion $|z| > 2$, at which point the trajectory diverges to infinity (cf. **Douady & Hubbard, 1982; Milnor, 2006**). The activation function is defined as:

$$f_M(x) = \frac{T(c)}{T_{max}} \tag{6}$$

where $c = \frac{x}{s}$ ensures the input range $[-1,1]$ maps to the region around the Mandelbrot set boundary. Scaling factor $s = 2.0$ focuses on the boundary $|z| = 2$, the established escape criterion. If no escape occurs within $T_{max} = 20$ iterations, we set $T(c) = T_{max}$. This ensures that points within the Mandelbrot set (which remain bounded indefinitely) receive maximum activation value 1, while fast-escaping points approach 0.

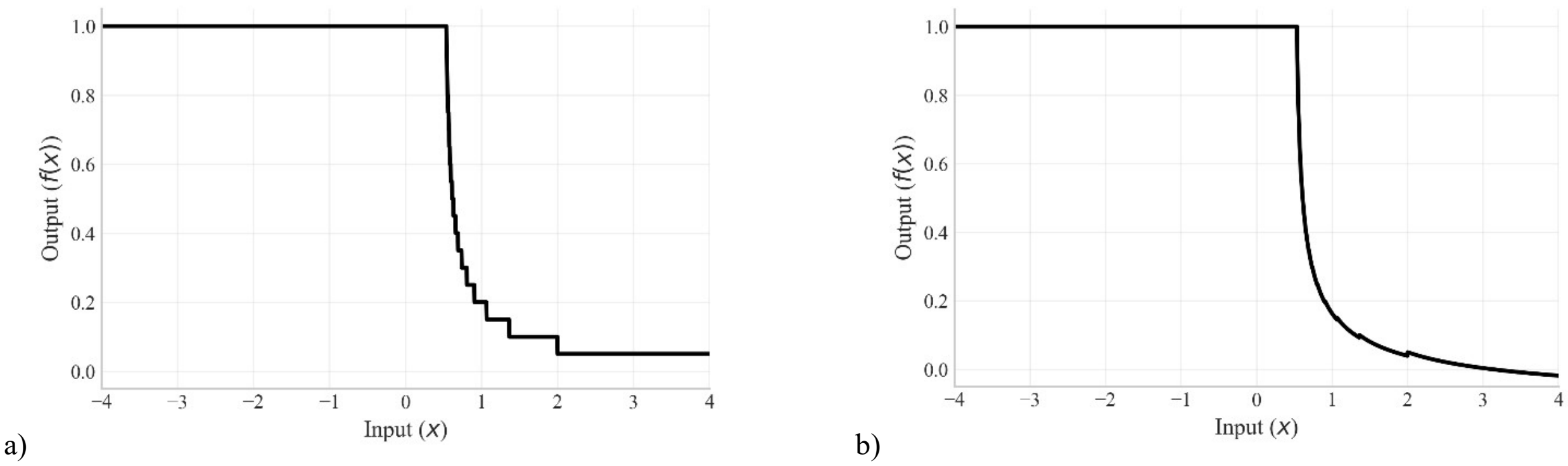

**Figure 1. Mandelbrot-based activation functions.** (a) Discrete variant using integer escape times, producing quantized output levels corresponding to iteration counts before divergence. (b) Continuous variant using smooth interpolation of escape times, providing infinite resolution within the unit interval. Both functions map the Mandelbrot set boundary region to high activation values (near 1.0) while fast-escaping points approach zero.

For the discrete variant (see **Fig. 1a**), $T(c) = \min\{n: |z_n| > 2\}$ if escape occurs within $T_{max} = 20$ iterations, otherwise $T(c) = T_{max}$.

For the continuous variant (see **Fig. 1b**), $T(c)$ uses smooth interpolation:

$T(c) = n - \log_2(\log_2(|z_n|))$ when $|z_n| > 2$, providing fractional escape times.

Both variants map $\mathbb{R} \to [0,1]$ through the normalization by $T_{max}$, with the discrete version producing at most 21 distinct output levels while the continuous version provides infinite resolution within the unit interval.

*3.2 Logistic Map-Based Activation Functions*

The logistic map activation functions, based on the chaotic dynamics studied by **May (1976)** and **Feigenbaum (1978),** can be represented as:

$$f_L(x) = ry(1 - \mathrm{y}), \tag{7}$$

where $r = 3.7$ (empirically chosen because it is within the chaotic regime $r > 3.57$ while being computationally stable), and $y$ is obtained through two variants:

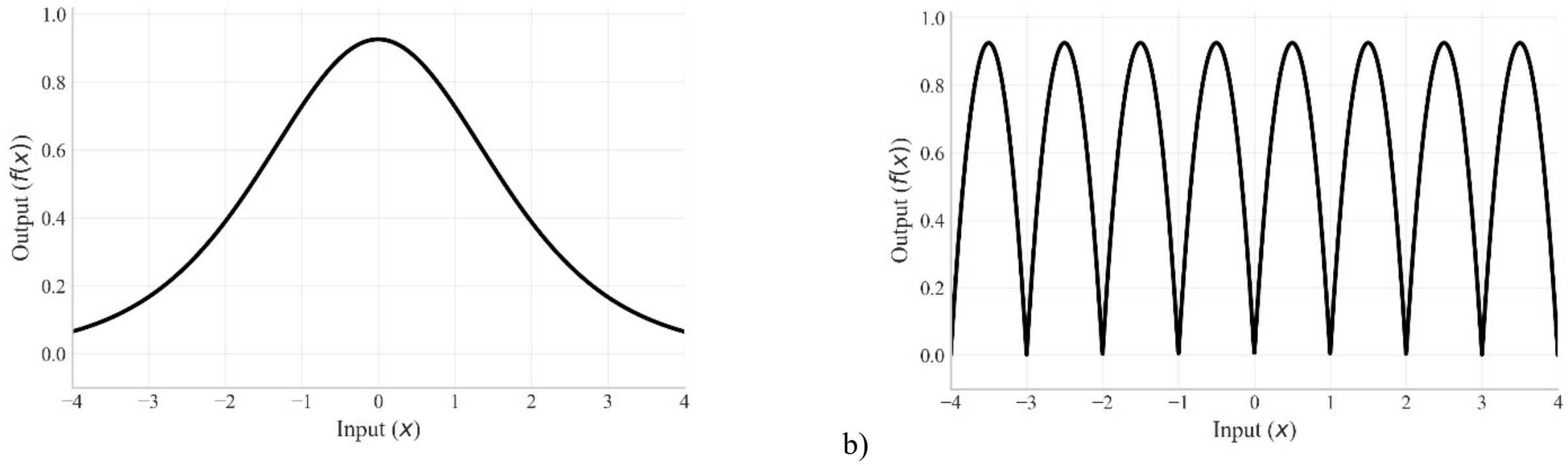

a) b)

**Figure 2. Logistic map-based activation functions with $r = 3.7$ (chaotic regime).** (a) Sigmoid variant applying chaotic logistic dynamics to sigmoid-transformed input, exhibiting sensitive dependence on initial conditions despite smooth appearance. (b) Modulo variant using fractional parts of input magnitude, creating periodic discontinuous jumps that preserve chaotic structure while introducing sharp transitions at integer boundaries.

For the sigmoid (smooth) variant (see **Fig. 2a**): $y = \sigma(x) = \frac{1}{1+e^{-x}}$.

For the modulo (discontinuous) variant (see **Fig. 2b**): $y = clip(|x| \; mod \; 1, \varepsilon, 1 - \varepsilon)$, where $|x| \; mod \; 1$ denotes the fractional part of $|x|$ (i.e., $|x| - \lfloor |x| \rfloor$), and $\varepsilon = 10^{-10}$ provides numerical bounds to prevent collapse. More formally, for the modulo variant:

$y = \max(\varepsilon, min(1 - \varepsilon, |x| \; mod \; 1))$

The sigmoid variant ensures smooth gradients but applies double nonlinearity, while the modulo variant introduces discontinuities that may enhance reservoir diversity through abrupt state transitions.

### *3.3 Weierstrass Activation Function*

The Weierstrass activation function, following the classical construction of a continuous but nowhere differentiable function **(Weierstrass, 1872; Hardy, 1916),** is defined as **(Fig. 3)**:

$$f_W(x) = \frac{1}{1-\mathrm{a}} \sum_{\kappa=0}^{\mathrm{K}-1} a^{\kappa} \cos(b^{\kappa} \pi s x) \tag{8}$$

where $a = 0.5$ to ensure rapid amplitude decay while maintaining sufficient fractal complexity. (Values closer to 1 would converge too slowly, and values closer to 0 would lose fractal characteristics). $b = 3$ provides adequate frequency scaling and ensures $ab > 1$. $K = 10$ balances mathematical accuracy with computational efficiency, normalized by $\frac{1}{1-a}$. The resulting function exhibits characteristic fractal spikes at multiple scales, demonstrating continuous but nowhere-differentiable behavior (see **Fig. 3**).

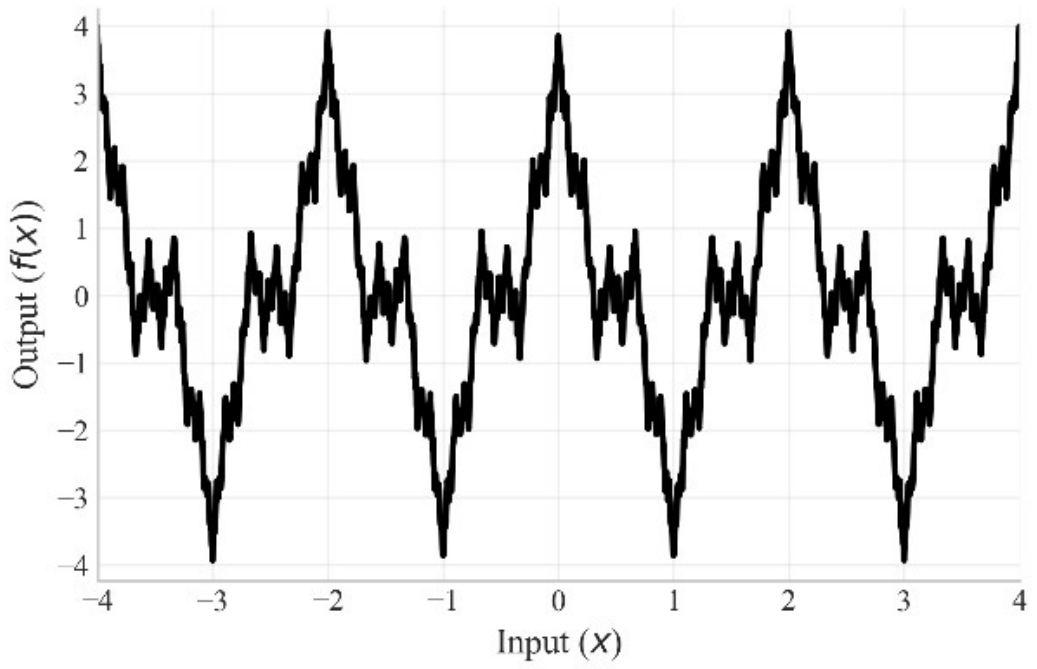

**Figure 3. Weierstrass activation function with $\boldsymbol{a = 0.5, b = 3, K = 10}$.** The function exhibits characteristic fractal behavior with self-similar spikes at multiple scales, demonstrating continuous but nowhere differentiable structure. The periodic oscillations with decreasing amplitude create the classic "pathological" function that challenged 19th-century mathematical intuitions about continuity.

### *3.4 Cantor-Based Activation Functions*

The Cantor function and set (e.g. **Cantor, 1883; Dovgoshey et al., 2006**), are both tested. The Cantor set is a fractal with the property of being uncountably infinite yet having zero Lebesgue measure, while the Cantor function (also known as the Devil's Staircase) is continuous everywhere but has zero derivative almost everywhere, rising from 0 to 1 despite being flat on intervals that collectively span almost the entire domain (**Salem, 1943**).

### 3.4.1 The Cantor Function (Devil's Staircase) Activation.

Given input $x \in \mathbb{R}$, the Cantor activation function $f_C(x)$ with depth parameter $d$ is:

$$f_C(x) = c_d(\sigma(x)) \tag{9}$$

where $\sigma$ maps $\mathbb{R} \to (0,1)$, $\sigma(x) = \frac{1}{1+e^{-x}}$ is the sigmoid function, and $c_d: [0,1] \to [0,1]$ is defined recursively:

$$c_d(y) = \begin{cases} y, & if\ d = 0 \\ \left(\frac{1}{2}\right) \cdot c_{d-1}(3y), & if\ y \in \left[0, \frac{1}{3}\right] \\ \frac{1}{2}, & if\ y \in \left(\frac{1}{3}, \frac{2}{3}\right) \\ \frac{1}{2} + \left(\frac{1}{2} \cdot c_{d-1}(3y-2)\right), & if\ y \in \left[\frac{2}{3}, 1\right] \end{cases}$$

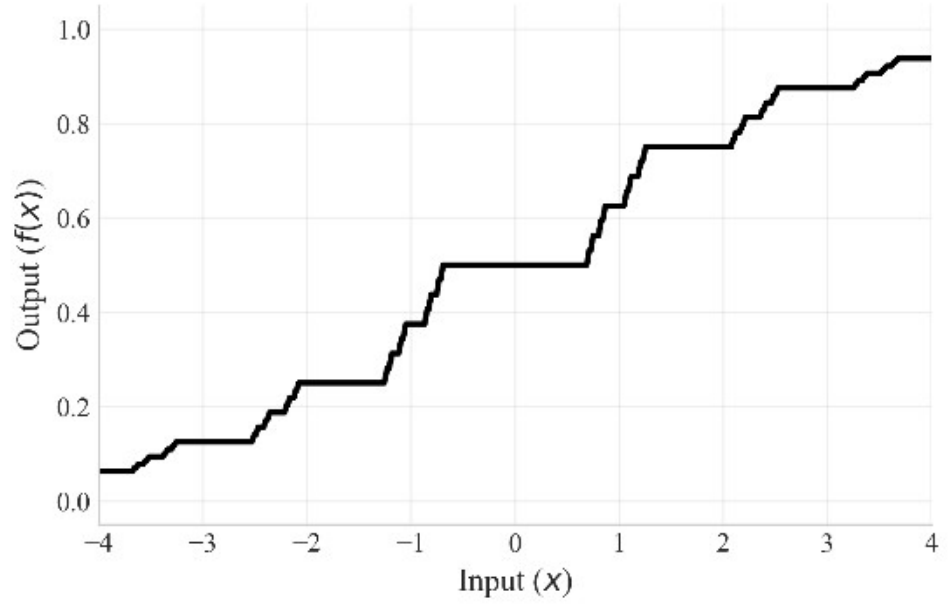

**Figure 4. Cantor function (Devil's Staircase) activation with depth $d = 10$.** The function exhibits monotonic increase from 0 to 1 while maintaining zero derivative almost everywhere, creating characteristic plateaus on middle-third intervals. Despite being continuous everywhere, the function demonstrates the counterintuitive property of rising through a set of measure zero.

The recursive construction creates the characteristic "devil's staircase," a monotonically increasing function that rises from 0 to 1 while having zero derivative almost everywhere (see **Fig. 4**). The function exhibits self-similar plateaus at multiple scales, remaining constant on the middle-third intervals removed.

### 3.4.2 The Cantor Set Activation.

The Cantor set activation uses an indicator function based on the ternary Cantor set. The activation $f_S(x)$ with depth $d = 10$ is defined as:

$$f_S(x) = 1_{\mathcal{C}_d}(\sigma(x)) \tag{10}$$

where $\mathcal{C}_d = \{y \in [0,1] : \text{for all } k \in \{0, \dots, d-1\}, FLOOR(3^k y\ mod\ 3) \neq 1\}$.

This constructs the classic ternary Cantor set by iteratively removing the middle thirds. This binary indicator function produces discontinuous jumps between 0 and 1, creating a sparse pattern of vertical spikes corresponding to points that survive the iterative middle-third removal process (see **Fig. 5**). Unlike the continuous Cantor function, this activation exhibits complete discontinuity with no intermediate values.

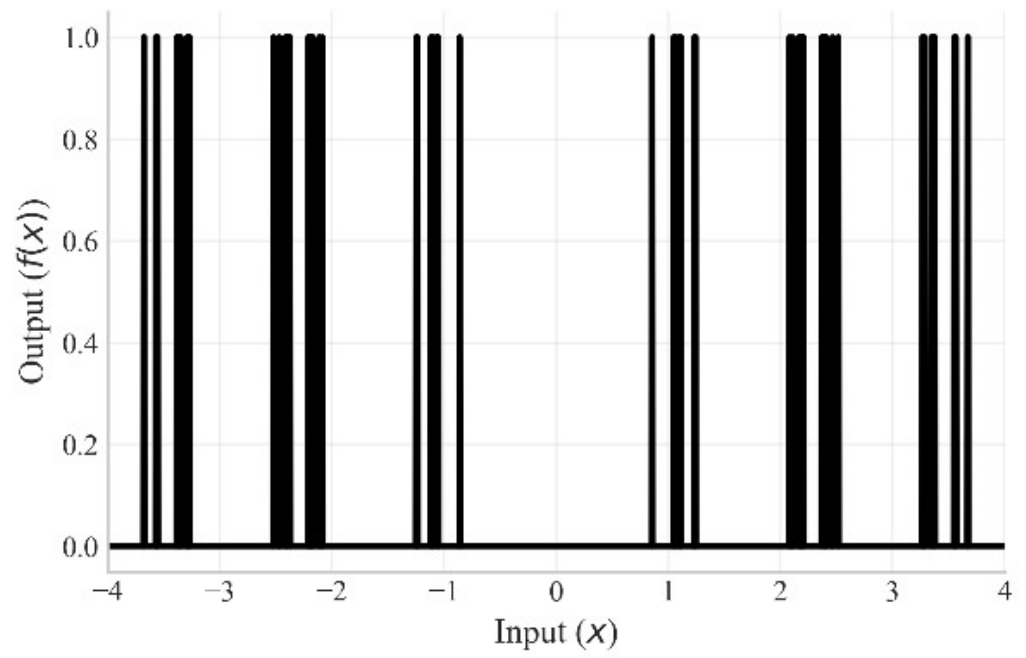

**Figure 5. Cantor set activation function with depth $d = 10$**. The binary indicator function produces discontinuous jumps between 0 and 1, creating sparse vertical spikes corresponding to points surviving the iterative middle-third removal process. The fractal distribution demonstrates uncountable infinity within zero measure - infinitely many activation points concentrated in a set of measure zero.

### *3.5. Brownian Motion Activation*

Adapted from **Wiener (1923),** our Brownian motion activation function introduces controlled stochastic perturbations at each state update (Note: this sacrifices reproducibility for the sake of testing stochasticity as an activation). Unlike input noise which affects only the input mapping, activation stochasticity fundamentally alters the state transition dynamics.

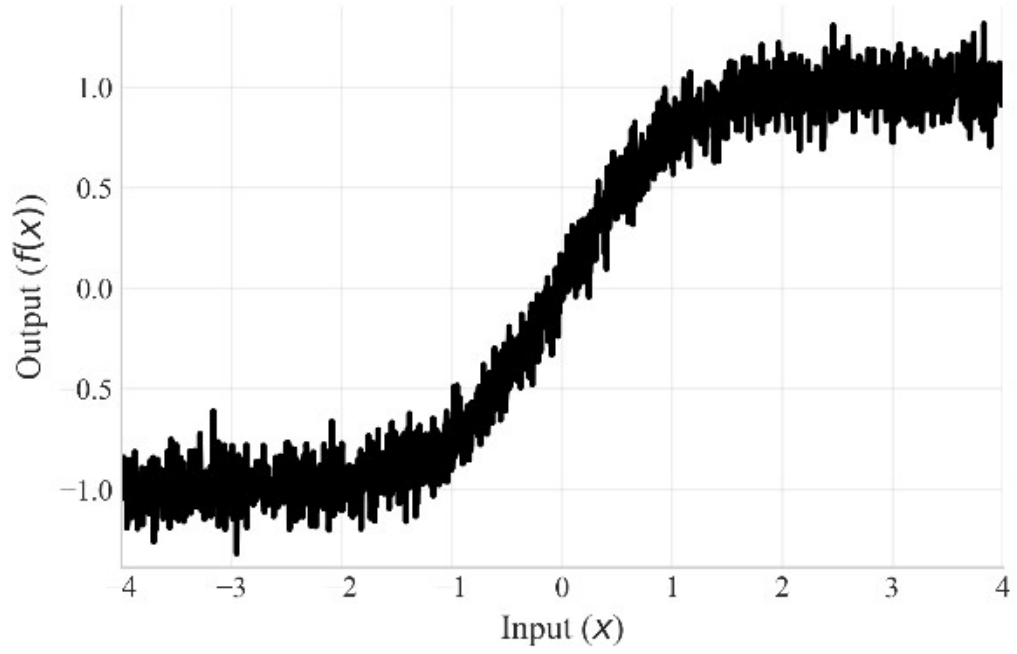

**Figure 6. Brownian motion activation function with $\eta = 0.3, \Delta t = 0.01$**. The function combines tanh nonlinearity with Gaussian noise increments, creating stochastic perturbations around the underlying sigmoid structure. Each evaluation produces different outputs for identical inputs, introducing genuine randomness into reservoir dynamics.

The Brownian motion activation function $f_B(x)$ with time step $\Delta t$ and scale $\eta$ is represented as **(Fig. 6)**:

$$f_B(\mathrm{x}) = \eta \cdot (tanh(\mathrm{x}) + \mathrm{W}_{\Delta \mathrm{t}}) \tag{11}$$

where $W_{\Delta t} \sim \mathcal{N}(0, \Delta t)$ is a Brownian increment drawn independently at each evaluation.

This introduces genuine stochasticity into the activation function, making each forward pass non-deterministic.

*Table 3.6. Activation Properties*

| Activation Function | Continuity | Differentiability | Monotonicity | Lipschitz | Empirical $L_{max}$ |
|---|---|---|---|---|---|
| | | ***Traditional Functions*** | | | |
| **Tanh** | Continuous | Yes | Increasing | Yes | 1.0 |
| **ReLU** | Continuous | Almost Everywhere | Increasing | Yes | 1.0 |
| | | ***Irregular Functions*** | | | |
| **Cantor Function (d=10) as implemented** | Continuous | Globally Lipschitz | Increasing | Yes | ~13† |
| **Cantor Set** | Discontinuous | Finite union of intervals | None | No | N/A |
| **Weierstrass (K=10) as implemented** | Continuous | Finite cosine sum | None | No | >500 |
| **Mandelbrot Continuous** | Continuous | Nowhere | None | No | ~16 |
| **Mandelbrot Discrete** | Discontinuous | Nowhere | None | No | ~0‡ |
| **Logistic Map (Sigmoid)** | Continuous | Yes | None | Yes | ~0.93 |
| **Logistic Map (Modulo)** | Discontinuous | Nowhere | None | No | N/A |
| | | ***Stochastic*** | | | |
| **Brownian Motion** | Discontinuous | Nowhere | None | No | N/A |

* Not globally Lipschitz, but Hölder continuous ($\alpha \approx 0.631$)† Non-Lipschitz behavior confined to measure-zero set; median local Lipschitz = 0

‡ Piecewise constant (L=0 almost everywhere) with discontinuous jumps at quantization boundaries

† Empirical Lipschitz constants estimated via finite differences ($\varepsilon = 10^{-6}, N = 100{,}000$ samples). Functions with $L_{max} < 20$ maintain ESP at large scale ($N = 2000$), while $L_{max} > 100$ correlates with ESP failure. Piecewise constant functions (Mandelbrot Discrete) exhibit $L \approx 0$ almost everywhere, and failures are due to discontinuous jumps at quantization boundaries.

*3.7 Computational Considerations*

Despite running up to 20 iterations per neuron per timestep, fractal activations remain computationally tractable. Our complete experimental suite (36,610 reservoir configurations across all conditions) completed in approximately 2 hours on a high-performance workstation (Intel i9, 128 GB RAM, NVIDIA RTX 4000 Ada, though GPU acceleration was not utilized).

Even on more modest hardware, the computational overhead remains manageable:

- The iterative computations are parallel and vectorize well
- Total runtime scales linearly with iteration count ($20 \times \text{iterations} \approx 1.7 \times \text{total runtime}$ in practice)
- No backpropagation required (this is the advantage of using reservoir computing)
- Memory requirements remain modest (largest reservoir: $N = 2000$)

For comparison, training modern deep learning models routinely requires days or weeks on similar hardware. The ability to explore extreme parameter regimes with fractal activations in hours rather than days makes them viable for rapid prototyping and deployment.

The 2.6x convergence speedup partially offsets the per-timestep cost. For real-time embedded systems with strict latency requirements, practitioners should benchmark on target hardware. However, for typical RC applications (signal processing, time-series prediction, classification), operating on millisecond or greater timescales, the overhead is acceptable given the stability benefits and convergence improvements.

## 4. Boundedness of Non-Smooth Activation Functions

While **Proposition 2.3** establishes that any bounded activation prevents state explosion, we prove boundedness for each of our non-smooth functions. Boundedness may seem trivial for some functions, but others (particularly Weierstrass as an infinite series and the Brownian variant with its stochastic component) require careful analysis to ensure reservoir weights are scaled appropriately.

**Proposition 4.1 The Mandelbrot activation functions are bounded.**

$f_M: \mathbb{R} \to [0,1]$ is bounded for both discrete and continuous variants.

**Proof:** For any $x \in \mathbb{R}$, we map $c = \frac{x}{s} + 0i$. The escape time $T(c) = \min\{n \in \{0, \dots T_{max}\} : |Z_n| > 2\}$ with $T(c) = T_{max}$ if no escape occurs. The discrete variant directly outputs $\frac{T(c)}{T_{max}} \in [0,1]$. The continuous variant uses smooth interpolation $T(c) = n - \log2(\log2(|z_n|))$ for escaping points, which still satisfies $T(c) \in \{0, 1, \dots, T_{max}\}$, since the log terms are bounded. Therefore, both variants map to $[0,1]$. ■

**Proposition 4.2 The logistic map activation functions are bounded.**

$f_L : \mathbb{R} \to [0, \frac{r}{4}]$ is bounded for both versions (sigmoid-wrapped and modulo-wrapped). For $r \in (0,4]$ and $f_L(x) = ry \cdot (1 - y)$, we define two preprocessors:

- $h_\sigma(x) = \sigma(x) = \frac{1}{1+e^{-x}}$ (sigmoid wrapper)
- $h_{mod}(x) = clip(|x| \; mod \; 1, \varepsilon, 1 - \varepsilon)$ where $\varepsilon = 10^{-10}$ (modulo wrapper) chosen to avoid instability while maintaining near-unit range.

Both satisfy $h(x) \in (0,1)$ for all $x \in \mathbb{R}$. For clarity, we denote the composed function as $g_h(x) = f_L(h(x))$ with $h \in \{h_\sigma, h_{mod}\}$:

$$0 < g_h(x) < \frac{r}{4} \text{ for all } x \in \mathbb{R}$$

**Proof:** The function $f_L(y) = ry(1 - y)$ on the domain (0,1) can be rewritten as:

$$f_L(y) = r\left[\frac{1}{4} - \left(y - \frac{1}{2}\right)^2\right]$$

Since $-\left(y - \frac{1}{2}\right)^2 \leq 0$ with equality only at $y = \frac{1}{2}$, we have:

$$f_L(y) \leq \frac{r}{4}$$

The maximum $\frac{r}{4}$ is achieved at $y = \frac{1}{2}$.

For the boundaries: as $y \to 0^+$ or $y \to 1^-$, we have $f_L(y) \to 0$.

Since both $h_\sigma$(x) and $h_{mod}(x)$ map into the open interval $(0,1)$ (never reaching the endpoints due to the sigmoid's asymptotic behavior and the clipping respectively), we have:

$$0 < f_l(h(x)) < \frac{r}{4}$$

For $r = 3.7$ specifically (for a computationally stable point in the chaotic regime): $g_h(x) \in (0, 0.925)$. ■

**Proposition 4.3 The Weierstrass function is bounded.**

$f_W : \mathbb{R} \to [-C, C]$ is bounded for some constant, $C$.

**Proof:** $f_W(x) = \frac{1}{1-a}\sum_{\kappa=0}^{K-1} a^{\kappa}\cos(b^k\pi sx)$. Since $|\cos(\theta)| \leq 1$ for all $\theta \in \mathbb{R}$, $|f_W(x)| \leq \frac{1}{1-a}\sum_{\kappa=0}^{K-1} a^{\kappa}|\cos(b^k\pi sx)| \leq \frac{1}{1-a}\sum_{\kappa=0}^{K-1} a^{\kappa}$. The geometric series sums to $\sum_{\kappa=0}^{K-1} a^{\kappa} = \frac{1-a^K}{1-a}$. Multiplying by the outer factor $\frac{1}{1-a}$ gives $|f_W(x)| \leq \frac{1-a^K}{(1-a)^2}$. If we check with variables $a = 0.5$ and $K = 10$: $|f_W(x)| \leq \frac{1-0.5^{10}}{(1-0.5)^2} \approx 3.996$. Therefore $f_W(x) \in [-3.996, 3.996]$ for all $x \in \mathbb{R}$. ■

**Proposition 4.4 The Cantor function is bounded.**

$f_C : \mathbb{R} \to [0,1]$ is bounded.

**Proof:** For any $x \in \mathbb{R}$, we first apply the sigmoid function $\sigma(x) = \frac{1}{1+e^{-x}} \in (0,1)$ for all $x \in \mathbb{R}$. The function $f_C(x) = c_d(\sigma(x))$ where $c_d$: $[0,1] \to [0,1]$. We prove by induction that $c_d(y) \in [0,1]$ for all $y \in [0,1]$:

Base case: When $d = 0, c_d(y) = y \in [0,1]$ by assumption. Inductive step: Assume $c_{d-1}(y) \in [0,1]$ for all $y \in [0,1]$. For any $y \in [0,1]$:

- If $y \in \left[0, \frac{1}{3}\right]: c_d(y) = \frac{1}{2} \cdot c_{d-1}(3y) \in \left[0, \frac{1}{2}\right]$ since $c_{d-1}(3y) \in [0,1]$
- If $y \in \left(\frac{1}{3}, \frac{2}{3}\right): c_d(y) = \frac{1}{2} \in [0,1]$
- If $y \in \left[\frac{2}{3}, 1\right]: c_d(y) = \frac{1}{2} + \left(\frac{1}{2}\right) c_{d-1}(3y-2) \in \left[\frac{1}{2}, 1\right]$ since $c_{d-1}(3y-2) \in [0,1]$

Therefore $c_d(y) \in [0,1]$ for all $y \in [0,1]$.

Since $\sigma(x) \in (0,1)$, we have $f_C(x) \in [0,1]$ for all $x \in \mathbb{R}$. ■

**Proposition 4.5 The Cantor set activation function is bounded.**

$f_S : \mathbb{R} \to \{0,1\}$ is bounded.

**Proof:** The function $f_S(x) = 1_{C_d}\sigma(x)$ is an indicator function that returns either 0 or 1. By definition, an indicator function $1_A(y)$ equals 1 if $y \in A$ and 0 if $y \notin A$. Since the range consists of only two values {0,1}, we have $f_S(x) \in [0,1]$ for all $x \in \mathbb{R}$. Therefore, $f_S$ is bounded with $|f_S(x)| \leq 1$. ■

**Proposition 4.6 The Brownian motion activation function is bounded.**

$f_B : \mathbb{R} \rightarrow [-\eta(1 + M), \eta(1 + M)$ is bounded in probability for finite $M$.

**Proof:** The function $f_B(x) = \eta(\tanh(x) + W_{\Delta t})$ where $W_{\Delta t} \sim N(0, \Delta t)$. Since $\tanh(x) \in [-1,1]$ for all $x \in \mathbb{R}$, we have:

$f_B(x) = \eta(\tanh(x) + W_{\Delta t})$ where $\tanh(x) \in [-1,1]$ and $W_{\Delta t}$ is a Gaussian random variable. For practical implementation with finite precision, we truncate the Gaussian at some multiple of standard deviations. Let $M = k\sqrt(\Delta t)$ for $k$ standard deviations (typically $k = 3$ or $k = 4$). Then with probability greater than $1 - 2\varphi(-k)$ (where $\varphi$ is the standard normal CDF): $|W_{\Delta t}| \leq M$.

Therefore, with high probability: $|f_B(x)| \leq \eta(|\tanh(x)| + |W_{\Delta t}|) \leq \eta(1 + M)$.

For $k = 3$, this holds with probability $> 0.997$. Thus, $f_B(x) \in [-\eta(1 + M), \eta(1 + M)$ with probability $> 0.997$. ■

**Remark.** While $f_B$ is not strictly bounded due to the Gaussian tail, it is effectively bounded for all practical purposes in finite-precision computation where extreme values are truncated.

## 5. Experimental Design

We evaluate ESP compliance of each activation function by testing whether two trajectories initialized from different states converge under identical inputs. Convergence is defined as $||x(t) - x'(t)|| < 0.1$ within 200 timesteps, where $x(t)$ and $x'(t)$ are reservoir states from different initial conditions. Functions failing to meet this criterion were evaluated for an extended 2,000 timesteps to distinguish slow convergence from genuine ESP violation. Complete scaling results across all network sizes are presented in **Appendix A.1.**

For configurations that satisfy ESP, we additionally measure convergence rate by computing the time at which $||x(t) - x'(t)||$ first falls below the threshold of 0.1, providing direct comparison of transient dynamics across activation functions.

*5.1 Convergence Testing Methodology*

Our work tests convergence from two initial conditions: zero state and random initialization with $||x(0)||_\infty = 2.0$, using Gaussian, uniform, and sparse input distributions. This follows **Yildiz et al. (2012)**, who proved through bifurcation analysis that pathological input distributions occur with a probability distribution of zero, justifying our focus on realistic input classes.

We evaluate ten activation functions across seven reservoir sizes ($N \in \{1, 10, 50, 100, 500, 1000, 2000\}$) with 1000 independent trials per configuration to ensure statistical robustness. Since results remain qualitatively similar across Gaussian ($\mu = 0$, $\sigma = 1$), uniform [-1,1], and sparse (90% zeros) input distributions, and having established boundedness for all our non-smooth functions, we can now apply **Proposition 2.3** to guarantee that reservoir states remain finite regardless of spectral radius.

*5.2 Parameter Sweep Methodology*

While boundedness prevents state explosion, it does not guarantee ESP-consistent convergence. To determine which non-smooth activations maintain ESP and at what parameter regimes, we conducted comprehensive parameter sweeps across reservoir configurations. To establish the operational boundaries of non-smooth activations beyond the baseline configuration ($\rho = 0.95, a = 0.7$), we conducted comprehensive parameter sweeps across $\rho \in \{0.5, 0.6, 0.7, 0.8, 0.9, \ 1.0, 2.0, 3.0, 4.0, 5.0\}$ and extended testing up to $\rho = 100$ for robust functions. For these tests, $a \in \{0.1, 0.3, 0.5, 0.7, 0.9\}$ with $N = 100$ neurons and 50 trials per configuration with five different random seeds.

## 6. Results

We present results from comprehensive parameter sweeps testing ESP compliance across activation functions, spectral radii, leak rates, reservoir sizes, and random seeds. We examine parameter sensitivity, revealing critical transitions in stability as $\rho$ and $a$ vary, implying that preprocessing topology and quantization effects, rather than boundedness alone, determine stability.

*6.1 Parameter Sensitivity Analysis*

**Figs 7 & 8** present the most interesting phase diagrams showing transitions in ESP compliance results across the $(\rho, a)$ parameter space for activation functions (1 if and only if all trials across all seeds converged). **Fig 9** shows spectral radius computation through eigenvalue analysis.

Key findings from the parameter sweep include:

1) **Spectral radius tolerance.** The logistic sigmoid **(Fig. 8a)** and Cantor function **(Fig. 8b)** remained stable throughout the grid (no failures up to $\rho \approx 5.0$ and $\rho \approx 10.0$ respectively, as shown in the extended parameter sweep), far exceeding the classical $\rho < 1$ heuristic for smooth activations (cf. **Jaeger, 2001**). This suggests bounded chaos can stabilize otherwise divergent dynamics, and it is consistent with a reduced effective gain from monotone compression/saturation, even though global worst-case bounds are $> 1$.

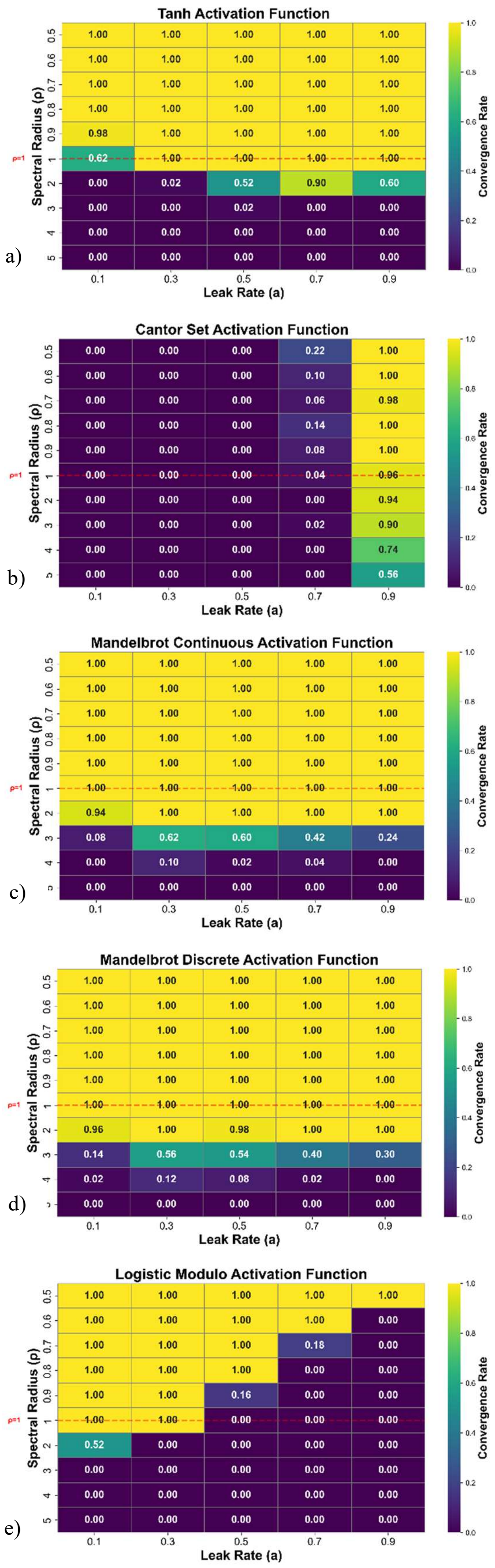

**Figure 7. Parameter sensitivity analysis of echo state property across non-smooth activation functions.** Heat maps show mean convergence rate across spectral radius $\rho$ and leak rate $a$ for $N = 50$. (a) Tanh baseline showing classical ESP behavior with sharp transition at $\rho \approx 1$. (b) Cantor set exhibiting binary stability requiring high leak rates ($a \geq 0.7$) and low spectral radii. (c) Mandelbrot continuous demonstrating moderate spectral radius tolerance up to $\rho \approx 3$. (d) Mandelbrot discrete showing similar behavior at small scales. (e) Logistic modulo displaying abrupt failure transitions sensitive to leak rate thresholds.

2) **Leak interactions.** Discontinuous, non-compressive activations (e.g. Cantor set, **Fig 7b**) require smaller $a$ for stability (robust region concentrated at $a \leq 0.5$). Whereas compressive maps (sigmoid wrapper, Cantor function) are largely insensitive to $a$ in this range.
3) **Critical thresholds.** All activations except the Cantor Function exhibit sharp phase transitions rather than gradual decay; empirical critical $\rho$ at $a = 0.7$ are approximately: tanh ~ 2.0, ReLU ~ 1.5, Mandelbrot continuous ~ 3.0 **(Fig. 7c),** Mandelbrot discrete ~ 2.0 **(Fig. 7d),** Cantor function $\rho \approx 10.0$ **(Fig. 8b),** logistic sigmoid $\rho \approx 5.0$ **(Fig. 8a).** (Full grids in **Appendix A.1**; boundaries shift mildly with $a$.)

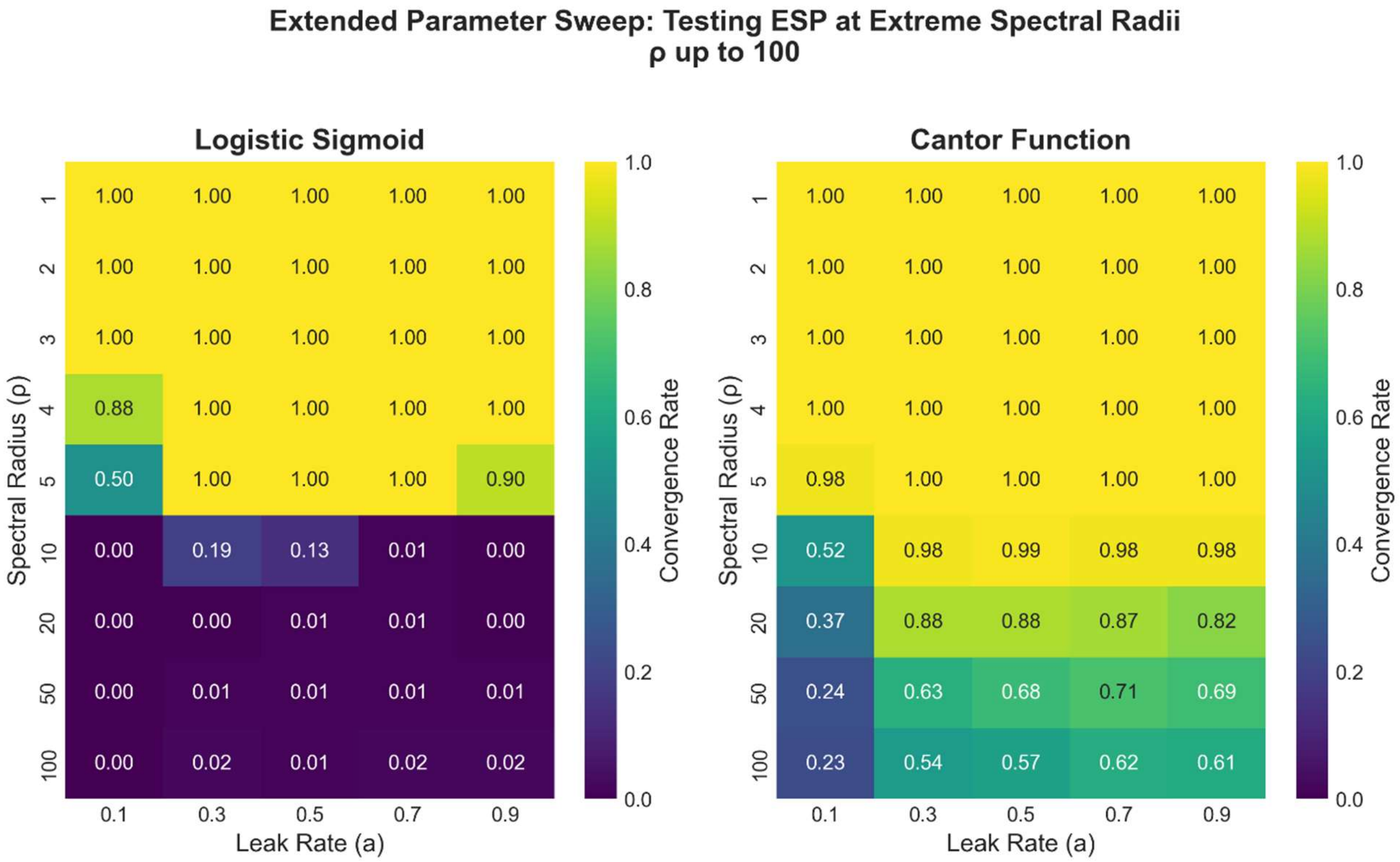

**Figure 8. Extended parameter sensitivity analysis of echo state property.** Extended parameter sweep testing ESP at extreme spectral radii up to $\rho = 100$. (a) Logistic sigmoid maintains ESP-consistent convergence up to $\rho \approx 5$, violating classical bounds by 5×. (b) Cantor function shows more gradual degradation, maintaining 50-60% convergence even at $\rho = 100$. Results demonstrate that specific activation functions can extend stable operating regions far beyond the traditional $\rho < 1$ constraint.

We confirmed that our reservoir weight matrices achieved the specified extreme spectral radii through direct eigenvalue computation by generating sparse random matrices and scaling them to achieve exact target spectral radii using the relation $W_{scaled} = W_{original} \times \frac{\rho_{target}}{\rho_{current}}$, where $\rho_{current} = \max(|\lambda|)$ from the eigenvalue decomposition. **Figure 9** confirms that our reservoirs achieve the specified spectral radii within numerical precision: $\rho = 10$ and $\rho = 100$ for $N = 500$ neurons. The eigenvalue distributions show that while most eigenvalues cluster near zero, the maximum eigenvalue magnitude precisely matches our target values, validating that the Cantor function's stability claims at $\rho \approx 10$ represent genuine mathematical behavior rather than computational artifacts.

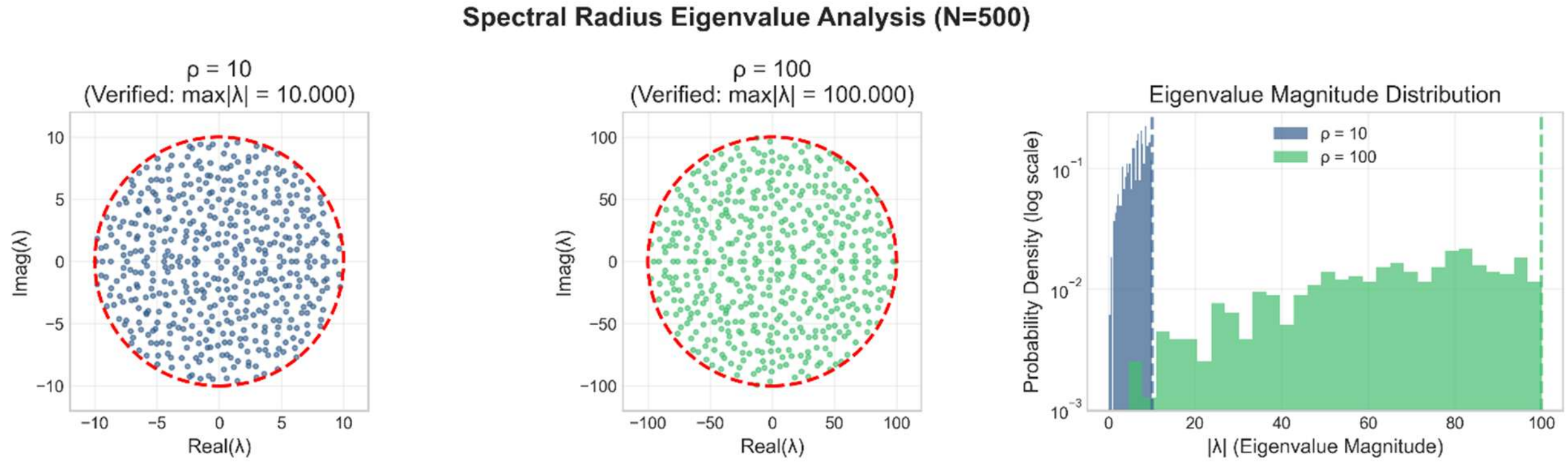

**Figure 9. Spectral radius verification through eigenvalue analysis**. Reservoir weight matrices ($N = 500$) constructed with target spectral radii of 10 and 100 show eigenvalues distributed exactly to the specified bounds. Despite these extreme spectral radii—far exceeding the classical $\rho < 1$ stability criterion—reservoirs with Cantor and logistic sigmoid activation functions maintain ESP-consistent convergence.

### *6.2 Scale-Dependent Convergence Behavior*

At $N = 1$, all activation functions maintained ESP as predicted by theory. However, scale-dependent failures emerged rapidly for non-smooth activations.

Our results reveal that several non-smooth activation functions not only maintain ESP at large scales, but also outperform traditional functions. The Cantor function achieved convergence in fewer timesteps than tanh and ReLU ($6.1 \pm 1.4$ vs $15.6 \pm 3.2$) at $N = 2000$, (as shown in **Figure 10**) while the continuous Mandelbrot and sigmoid-wrapped logistic map maintained perfect stability with convergence distances below $10^{-14}$. Most remarkably, these performance gains come from functions that violate the basic smoothness assumptions underlying most RC theory.

### *6.2.1 Early scale dynamics ($N = 10$ to $N = 500$)*

The Brownian motion activation showed immediate instability, achieving convergence in only 1.6% of trials at $N = 10$, consistent with ESP being defined for a deterministic reservoir map. Internal stochasticity violates those assumptions and typically disrupts fading-memory behavior (cf. **Gonon & Ortega, 2021**; noise can regularize inputs but is not equivalent to stochastic activations inside the map, **Wikner et al., 2024**).

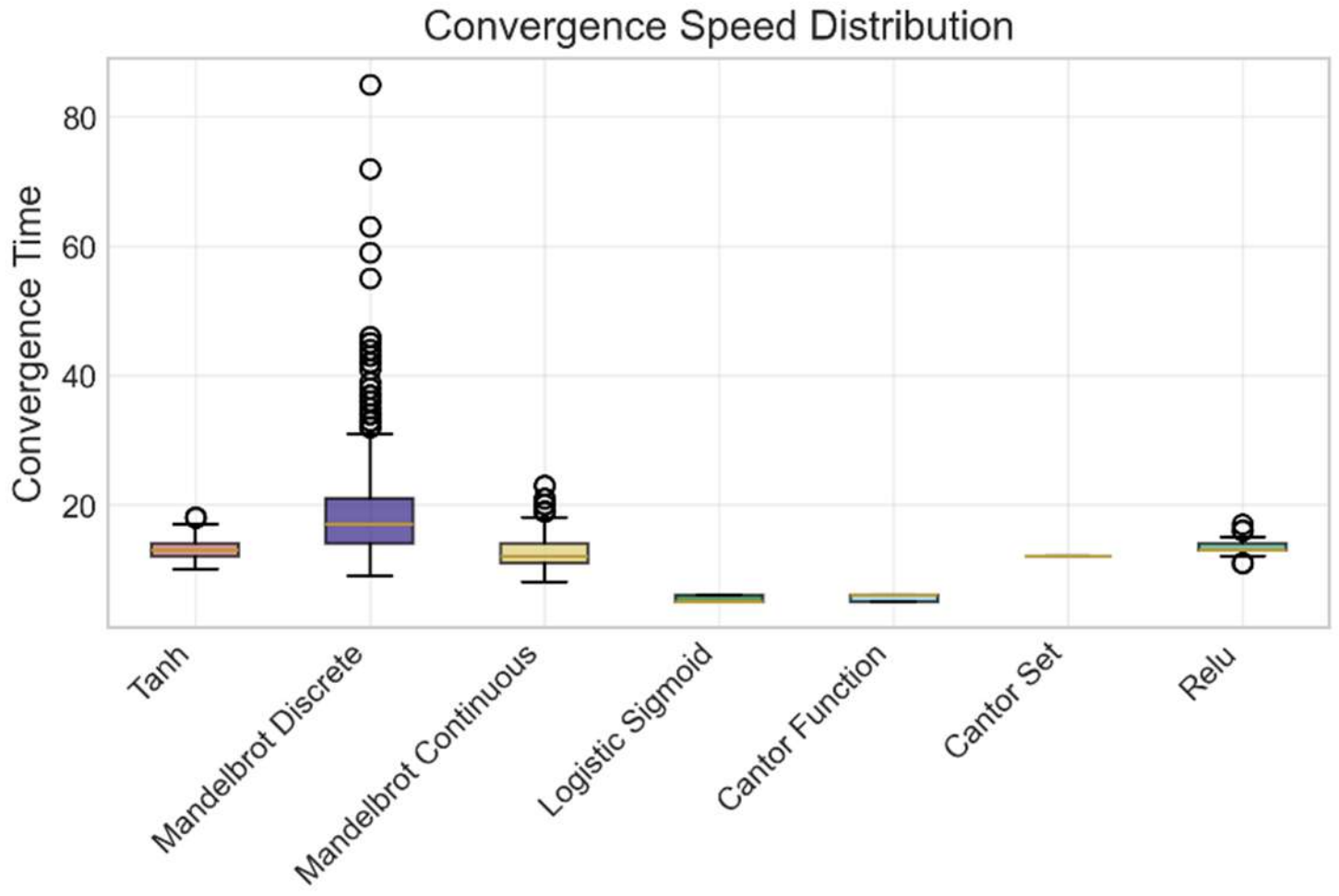

**Figure 10. Convergence speed distribution across activation functions.** Box plots show the number of timesteps required for state trajectories to converge (y-axis) under large-scale reservoir settings ($\rho = 0.95$, $a = 0.7$, $N = 500$). Classical smooth activations (tanh, ReLU) exhibit median convergence around 15 timesteps, with wide variance for Mandelbrot (discrete). In contrast, the logistic sigmoid and Cantor function converge nearly instantaneously (median ≈ 6 timesteps), achieving more than 2.5 times faster stabilization relative to tanh and ReLU. This snapshot illustrates the asymptotic regime; qualitatively similar patterns were observed across all N tested (1-2000, shown in the Appendix.)

The Weierstrass function demonstrated limited stability at $N = 10$ (28.5% convergence) before catastrophic failure at $N = 50$ (0.1% convergence), aligning with reports that reservoir state geometry can become increasingly fractal and irregular with scale (**Mayer & Obst, 2022**). The logistic map with modulo preprocessing exhibited a sharp phase transition between $N = 10$ ($\approx 100\%$ convergence) and $N = 50$ (convergence $< 1\%$).

The Cantor set activation displayed increasing instability with scale, achieving 70.6% convergence at $N = 100$ with significantly delayed convergence (mean: 75.3 timesteps, range: 20-200 timesteps) and marginal exponential decay rate ($\lambda \approx -0.05$), indicating proximity to ESP violation, consistent with input-relative ESP fragility near critical regimes (**Yildiz et al., 2012**). By $N = 500$, Cantor set convergence decreased to 0.4%.

The Mandelbrot (discrete and continuous), logistic with sigmoid preprocessing, and Cantor function maintained 100% convergence through $N = 100$. Notably, the discrete Mandelbrot exhibited convergence to a finite attractor (final distance: $0.033 \pm 0.012$) rather than point convergence, with extended convergence times (20-80 timesteps) at $N = 500$. The continuous Mandelbrot showed non-monotonic convergence behavior, with some trajectories exhibiting temporary increases in state distance before continuing to converge, creating a "peaking" appearance in the aggregate appearance, as shown in **Figure 11.** This echoes observations that reservoir dynamics can exhibit fractal structure and multi-scale transients (**Mayer & Obst, 2022**).

Of further note, we observed unexpected instability from ReLU at $N = 10$. It achieved 99.9% convergence with rare catastrophic failures resulting in numerical overflow (final distance $> 10^9$ in 0.1% of trials). This suggests that even the "safe," smooth activations can exhibit pathological edge cases under certain initialization conditions.

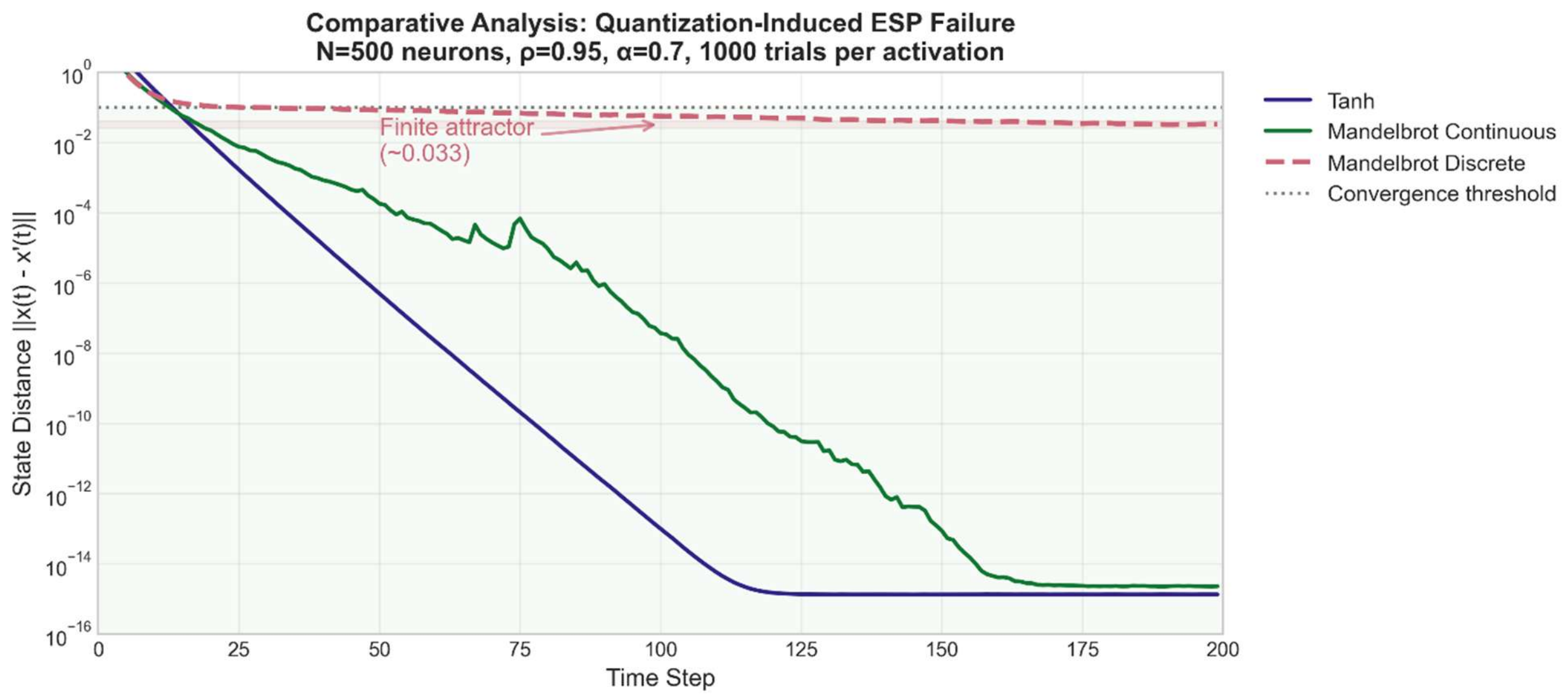

**Figure 11. Comparative analysis of echo state property (ESP) compliance across activation functions at scale ($\rho = 0.95$, $a = 0.7, N = 500$, 1000 trials per activation).** Mean state distance evolution highlights the anomalous behavior of the discrete Mandelbrot activation including its convergence to a finite attractor, along with the continuous Mandelbrot's metastable terraces and brief divergence peaks.

### *6.2.2 Critical scaling regime ($N = 1000$ to $N = 2000$)*

At $N = 1000$, a critical divergence emerged between the discrete and continuous variants of the Mandelbrot activation. The discrete Mandelbrot exhibited degraded performance (60.0% convergence, mean final distance: $0.1802 \pm 0.067$ when converged), while the continuous variant maintained perfect stability (100% convergence; final distance < $10^{-14}$). This divergence represents the onset of quantization-induced instability, with the idea that deterministic, continuous reservoir maps underpin fading-memory behavior (cf. **Gonon & Ortega, 2021**) and that small architectural changes can trigger sharp stability transitions in ESNs (cf. **Racca & Magri, 2021**). The quantization ratio (**Eq. 2**) reached 47.6 for $k = 21$ discrete levels.

Of note, the discrete Mandelbrot at $N = 1000$ exhibited an anomalous steady-state error of $\approx 10^{-1}$ rather than converging to zero, suggesting convergence to a spurious attractor rather than true ESP. This implies that quantization-induced failure manifests not as divergence but as convergence to an incorrect fixed point. This metastable state persisted across both the 200-steps and the 2000-steps horizons. Complete phase evaluation charts are available in **Appendix A.1.**

At $N = 2000$, the discrete Mandelbrot showed near-complete ESP failure (0.4% convergence; final distance $0.2817 \pm 0.134$ when converged), confirming the predicted critical threshold where $Q = 95$. In stark contrast, the continuous Mandelbrot maintained 100% ESP convergence (final distance: $0.0032 \pm 0.0018$), indicating that smooth interpolation of escape times eliminates the quantization bottleneck entirely (see also **Ozturk et al., 2007** on stability's dependence on input scaling/topology).

Among other activations at $N = 2000$, only tanh (100% convergence; $15.6 \pm 3.2$ timesteps), Cantor function (100% convergence; $6.1 \pm 1.4$ timesteps), ReLU (100% convergence; $15.2 \pm 2.8$ timesteps), and logistic map with sigmoid preprocessing (100% convergence; $6.0 \pm 1.2$ timesteps) remained stable.

Remarkably, the Cantor function converged in fewer timesteps than traditional activation functions despite being non-differentiable everywhere, suggesting that its piecewise-constant structure with infinite discontinuities paradoxically enhances convergence, potentially through dimensional reduction. Related work reports multiscale, fractal-like state geometry in reservoirs (**Mayer & Obst, 2022**) and that noise can regularize predictions when applied externally to the learning pipeline (**Wikner et al., 2024**).

Additionally, the logistic map with sigmoid preprocessing achieved 100% stability with final distance of 0.0000. Logistic map (sigmoid wrapper) and the Cantor function achieved mean final distances below $10^{-14}$ (numerically indistinguishable from zero in double precision), challenging conventional assumptions about regularity requirements for ESP and the gold-standard use of ReLU and tanh as activation functions for speed in RC applications.

In short, the $N = 2000$ results establish a clear hierarchy: continuous activations can maintain ESP-consistent behavior at scale, quantized functions are prone to failure at predictable thresholds, and stochastic/discontinuous preprocessing appears fundamentally incompatible with large-scale RC. **Fig 12** presents the scaling behavior of ESP convergence across network sizes.

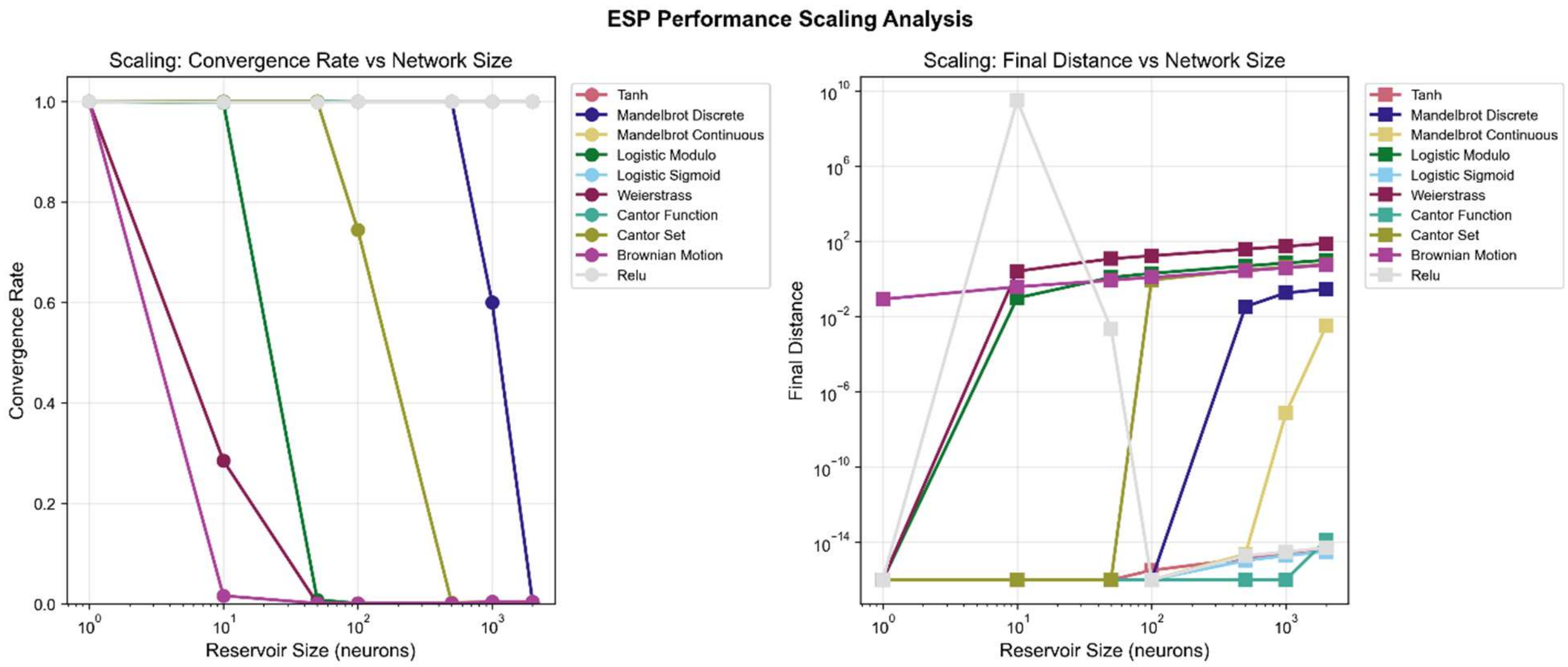

**Fig 12. Scaling analysis of ESP-consistent convergence across activation functions and reservoir dimensions.** (Left) Convergence fraction, defined as the proportion of 1000 trials per configuration where $||x(t) - x'(t)|| < 0.1$ within 200 timesteps, for each activation function across network sizes $N \in \{1, 10, 50, 100, 500, 1000, 2000\}$. Three distinct scaling behaviors emerge: scale-invariant functions maintaining 100% convergence (tanh, ReLU, Cantor function, Mandelbrot continuous, logistic sigmoid), functions exhibiting sharp phase transitions at critical sizes (discrete Mandelbrot at $N \approx 2000$, Cantor set at $N \approx 100$), and early-failure functions (logistic-modulo, Weierstrass, Brownian motion).
(Right) Mean final state distance after 200 timesteps (log scale). Stable activations achieve convergence to within $\approx 10^{-15}$, numerically indistinguishable from zero in double precision, while unstable cases diverge above $10^0$, a fifteen order-of-magnitude separation. The discrete Mandelbrot activation shows a sharp transition at $N = 2000$, consistent with the predicted critical quantization ratio of $Q = {}^{N}/_{k} \approx 95$ (See: **Eq. 2**) for $k = 21$ discrete levels. Total experiments: 36,610 reservoir configurations tested with spectral radius $\rho = 0.95$ and leak rate $a = 0.7$. The ReLU spike at $N = 10$ reflects rare catastrophic-overflow trials in an otherwise 99.9%-convergent regime (see Figure A.1.2 for detail); it is not representative of ReLU's typical behavior at this scale.

## 7. Mechanistic Analysis

Our results imply that preprocessing structure, not continuity per se, is the key determinant of whether an activation function supports ESP-consistent convergence. Further, our results imply that monotonicity, not smoothness, contributes most to the fading memory property.

*7.1 Sigmoid Versus Modulo Wrapping*

The divergent performance of logistic map variants provides critical insight into the role of preprocessing topology in reservoir stability. Both logistic map activations implement the same chaotic dynamics ($f_L(y) = ry(1-y)$ with $r = 3.7$, but differ in their preprocessing of the affine input.

**7.1.1 Sigmoid-Wrapped Logistic Map.**

With $g(x) = f_L(\sigma(x))$ and $\sigma(x) = \frac{1}{1+e^{-x}}$, the chain rule gives $|g'(x)| = |f'_L(\sigma(x))|\,|\sigma'(x)| \leq r \cdot \max_x |\sigma'(x)| = r \cdot \frac{1}{4} = 0.925$.

For the leaky ESN update $F(x) = (1-a)x + ag(W^{in}u + W^{res}x)$, a sufficient contraction condition is $Lip\ (F) \leq (1-a) + aL_g\left||W^{res}|\right| < 1$ for an induced operator norm $||\cdot||$ (e.g. $\ell_\infty$ or spectral). With $a = 0.7$, $L_g = 0.925$, and typical $\left||W^{res}|\right| \approx \rho = 0.95$, we obtain $(1-a) + aL_g\left||W^{res}|\right| \approx 0.915 < 1$, explaining the 100% ESP-consistent convergence we observe. (Sufficient-condition flavor per **Jaeger, 2001; Buehner & Young, 2006; Ozturk et al., 2007; Grigoryeva & Ortega, 2018**).

**7.1.2 Modulo-Wrapped Logistic Map.**

With $g(x) = f_L(m(x)), m(x) = |x|\ mod\ 1$ introduces countably many discontinuities at integer boundaries, so $g$ is not globally Lipschitz (no finite $L_g$). Consequently, $Lip(F)$ cannot be bounded by a number $< 1$ via the same route. Small changes in $x$ can map to distant points on [0, 1]. Empirically this yields the sharp loss of convergence we see once $N$ crosses the critical scale shown in **Section 6.2**. This matches the broader observation that stability is sensitive to preprocessing topology and effective gain (**Ozturk et al., 2007; Racca & Magri, 2021**).

These results dovetail with input-relative ESP (pathological sequences have probability zero) and with the fading-memory ⟺ ESP link: deterministic, continuous maps with sub-unit effective gain tend to preserve fading memory, while discontinuous preprocessing breaks it (**Yildiz et al., 2012; Grigoryeva & Ortega, 2018**).

*7.2 Discrete Versus Continuous Escape Time*

The Mandelbrot activation variants reveal how quantization affects scaling limits. Both compute the iteration $z_{n+1} = z_n^2 + c$, but differ in output resolution.

**7.2.1 Discrete (quantized) Mandelbrot escape time.**

$T_{disc}(c) \in \{0, 1, \dots, k-1\}$ with $k = 21$, induces a finite alphabet per neuron and partitions the reservoir state space into $k$ equivalence classes per coordinate.

**7.2.2 Continuous (smooth) Mandelbrot escape time.**

$T_{cont}(c) = n + 1 - \frac{\log(\log|z_n|)}{log2}$ for the first $n$ with $|z_n| > 2$ (any fixed bailout $> 2$ works up to constants). This yields a continuous, real-valued feature on [0,1] after normalization.

**7.2.3 Mathematical implications of Mandelbrot distinctions.**

The Mandelbrot variants demonstrate why quantization fails at scale. Let $f$ be any activation whose image has finite cardinality $k$ (k-level quantization). For a fixed input sequence, the leaky ESN update in **Eq. 1** induces a deterministic map on a finite set at most $k^N$ (post-activation states). Hence, every trajectory is eventually periodic (finite dynamical systems folklore), and multiple distinct cycles / spurious fixed points generally exist. ESP requires a unique asymptotic state sequence for a given input. The presence of multiple cycles means two initial conditions can converge to different attractors. Thus, ESP can fail even though all states remain bounded.

We observe that, at $N = 1000$ the crowding ratio (see **Eq. 2**) $Q = 47.6$. The discrete variant develops a steady-state error $\approx 10^{-1}$ rather than the point convergence, consistent with convergence to a spurious attractor in the finite-state dynamics as observed in the discrete Mandelbrot results. By $N = 2000$ ($Q \approx 95$) near-complete ESP failure occurs. The heuristic "crowding" grows with dimension: the effective map acts on $k^N$ bins. As $\frac{N}{k}$ rises, neighboring pre-activation states straddle bin boundaries, decreasing contraction.

The continuous variant, by contrast, remains deterministic and globally Lipschitz after normalization. With leak $a$ and $||W^{res}||$ in our regime, the leak-adjusted gain $(1 - a) +$ $aL_g||W^{res}|| < 1$ holds, giving ESP-consistent convergence even at $N = 2000$ with final distances $\lesssim 10^{-14}$.

**Remark.** This analysis suggests that quantized activation functions have a fundamental scaling limit approximately proportional to the number of quantization levels $k$. For practical ESN implementations, this implies that discrete activations may be suitable for small reservoirs but will inevitably fail as $N$ grows, regardless of other hyperparameters.

*7.3 Monotonicity*

We find that monotonicity is helpful but not required for ESP. The Cantor function (monotonically increasing) and the logistic sigmoid (wrapped with monotonic preprocessing) outperformed the other functions by maintaining ESP at the highest spectral radii. However, the continuous Mandelbrot (non-monotonic) maintained ESP consistently across leak rates and random seeds. The discrete Mandelbrot (non-monotonic) maintained ESP with respect to the ratio outlined in **Eq. 2.** The Weierstrass function, Cantor set, and Brownian motion (non-monotonic) did not achieve ESP either at all or consistently at large scale despite being bounded.

Thus, monotonic irregular functions demonstrate superior performance across all metrics: faster convergence, higher spectral radius tolerance, and scaling from $N = 1$ to $N = 2000$.

Our empirical results further reveal (as shown in the charts in the **Appendix**), a strong connection between monotonicity and fading memory in non-smooth activations. Monotonic smooth functions (ReLU, tanh) and monotonic non-smooth functions (Cantor function and logistic sigmoid) exhibit rapid exponential decay of state differences (median 6.0-15.6 timesteps to convergence), consistent fading memory behavior across all reservoir sizes ($N = 1$ to $N = 2000$), and preservation of temporal order: if $u_1(t) < u_2(t)$, then $\gamma(u_1(t)) \leq \gamma(u_2(t))$.

Mandelbrot discrete and Mandelbrot continuous (non-Monotonic, non-smooth) maintain ESP for at least some leak rates at $N = 2000$ and show fading memory connection with ESP as with the smooth functions. Whereas, the other non-monotonic non-smooth functions (Weierstrass, Cantor set, logistic modulo) fail to maintain ESP at $N \geq 10$, loss of fading memory as evidenced by non-converging trajectories, and unpredictable mapping of input sequences to reservoir states.

### 7.3.1 The Cantor Function has Monotonicity without smoothness.

The Cantor function provides a striking counterexample to the assumption that smooth derivatives are necessary for fading memory. Despite being continuous everywhere but derivative of 0 almost everywhere, non Lipschitz continuous (but Hölder continuous with exponent $\frac{log2}{log3} \approx 0.631$), and self-similar with fractal structure at all scales, the Cantor function achieves convergence in the fewest timesteps in our experiments (6.1 vs. 15.6 for tanh) and maintains ESP up to spectral radius $\rho \geq 10$. The mechanism appears to be its strict monotonicity combined with another property, and this warrants future exploration. On the middle-third intervals (which comprise almost the entire domain), the Cantor function is locally constant, effectively performing dimensional reduction while preserving order relationships.

### 7.3.2 Implications for Fading Memory Theory

Our results suggest that monotonicity, not smoothness, is the critical property for fading memory in bounded activation functions. This has several implications. Fading memory does not require differentiability. The Cantor function demonstrates that nowhere-differentiable functions can exhibit excellent fading memory properties.

Combined with the leak rate $a$, bounded monotonic activations empirically show more consistent contraction dynamics than non-monotonic ones (**Section 6.2**), though the precise mechanism linking monotonicity to fading memory in reservoirs with mixed-sign weights remains an open question. In contrast, non-monotonic functions can (but do not necessarily) map temporally adjacent inputs to distant activation values, destroying the causal structure necessary for fading memory (e.g. Weierstrass oscillates wildly at all scales). Additionally, monotonicity is not required for fading memory in bounded activation functions. The Mandelbrot functions maintained ESP and the fading memory property despite being non-monotonic and nowhere differentiable. This is in contrast to the Weierstrass function that never maintained ESP or fading memory property.

### *7.4 Empirical Lipschitz Analysis*

To understand why certain non-smooth activations maintain ESP while others fail, we estimated local Lipschitz constants by computing $|f(x+\varepsilon) - f(x)|/_{\varepsilon}$ for 100,000 randomly sampled points with $\varepsilon = 10^{-6}$. The continuous Mandelbrot activation exhibits a maximum local Lipschitz constant of approximately 16, with median value of 0.017, indicating it is globally Lipschitz continuous despite being nowhere differentiable in the classical sense.

With the leaky update **(Eq. 1)** and typical parameter values ($a = 0.7, \rho = 0.95$), the effective gain $(1 - a) + aL_M||W^{res}|| \approx 0.3 + 0.7(16)(0.95) \approx 10.9$, which exceeds 1. However, this worst-case bound is conservative. The 95th percentile local Lipschitz constant is only 0.30, suggesting that for typical inputs, the effective contraction is much stronger than the worst-case analysis predicts.

In contrast, the discrete Mandelbrot exhibits a maximum local Lipschitz constant of approximately 0, its piecewise constant almost everywhere. However, this does not indicate superior regularity. Rather, the function has discrete jumps at quantization boundaries (where escape time changes from iteration $n$ to $n + 1$). These discontinuities occur on measure-zero sets but become problematic as reservoir size N grows. As shown in **Section 7.2**, when the crowding ratio **(Eq. 2)** exceeds critical thresholds ($Q \approx 47$ at $N = 1000, Q \approx 95$ at $N = 2000$), the quantization-induced finite state dynamics admit multiple competing attractors, causing ESP failure even though states remain bounded.

In stark contrast, the Weierstrass function, while continuous and bounded, exhibits maximum local Lipschitz constant exceeding 545, with median value 176. This extreme local amplification of over 30x larger than the continuous Mandelbrot explains its complete failure to maintain ESP. The nowhere-differentiable structure creates arbitrarily steep local gradients that disrupt contraction even with the leak term.

The Cantor function presents an intermediate case: maximum local Lipschitz constant of approximately 13, but median value of 0.0, reflecting its piecewise-constant structure. The function is flat (derivative zero) on the middle-third intervals comprising measure $(\frac{2}{3})^d$ of the domain at depth $d$, with rapid transitions confined to a Cantor set of measure zero. While not globally Lipschitz, the non-Lipschitz behavior occurs on a negligible set, and the monotonicity constraint prevents these jumps from causing divergence.

These findings reveal that the global Lipschitz continuity with moderate constant ($L \leq 20$) is sufficient for ESP in bounded activations, while extreme local variation ($L > 100$) is incompatible with ESP regardless of other properties. The Cantor function demonstrates that bounded, monotonic functions can maintain ESP even when Lipschitz continuity fails, provided the violations are confined to measure-zero sets.

**Claim 7.5 Quantized activations induce eventual periodicity.**

For activation functions with finite image size $k$, the reservoir state space is partitioned into at most $k^N$ equivalence classes. Under any fixed input sequence, the dynamics become a finite deterministic map, which must eventually enter a periodic orbit. ESP holds if and only if this orbit is unique (independent of initial conditions) for each input sequence. In practice, for large $N$ and $k \ll N$, multiple orbits appear, yielding ESP-inconsistent behavior. This precisely explains the discrete Mandelbrot's failure at $N = 1000$, where the crowding ratio of $47.6$ (**Eq. 2**) exceeds the threshold for unique orbit convergence, while the continuous variant with infinite resolution ($k = \infty$) maintains ESP at all scales.

*7.6 Why Classical ESP Theory Is Too Conservative*

Our results refine the classical view: a global Lipschitz constant $< 1$ for the reservoir update remains a sufficient condition for ESP, but it is not necessary and can be overly conservative.

1. **Not necessary.** The Cantor function activation (continuous, singular, non-differentiable; Hölder exponent $\frac{log2}{lo}$ indicating it's smoother than a discontinuous function but rougher than a differentiable one) achieves ESP-consistent convergence in our parameter sweeps, showing that smoothness/Lipschitz continuity is not required due to preprocessing topology yielding a compressive, non-expansive map.
2. **Effective versus global bounds.** For the logistic map with sigmoid wrapper, the composed activation has $L_g = r \cdot \max|\sigma'| = 0.925$. With leak rate $a$ and $||W^{res}|| \approx \rho$, the leak-adjusted gain $(1 - a) + aL_g||W^{res}|| \approx 0.915 < 1$ explains the near-machine-precision convergence. Thus, it is the preprocessing-induced effective gain, not the chaotic activation function, that predicts stability (cf. **Manjunath & Jaeger, 2013**).
3. **Quantization and scaling.** Discrete activations (finite $k$ levels) can be stable at small $N$, but as the crowding ratio (see **Eq. 2**) grows, they admit multiple competing attractors, yielding ESP-inconsistent behavior (steady-state errors/limit cycles). Our data indicate practical thresholds in $Q$; formally ESP in the quantized case requires a unique global attractor for the induced finite dynamical system.

**Takeaway.** ESP analysis should be activation-specific: (i) check leak-adjusted effective gain for the composed map; (ii) account for preprocessing topology (compressive/monotone vs. dispersive/discontinuous); and (iii) for quantized maps, reason about attractor uniqueness rather than global Lipschitz bounds.

## 8. Discussion

Our findings provide new perspectives on the long-standing edge-of-chaos debate in reservoir computing (RC). While **Bertschinger and Natschläger (2004)** suggested optimal performance at the critical transition, and **Carroll (2020)** found this non-optimal for real applications, both studies relied exclusively on smooth activation functions. Our results suggest the debate may have been conflating two separate issues: the spectral radius operating point and the activation function geometry itself. Our results suggest that the apparent conflict between edge-of-chaos theory and practical applications may stem from an incomplete exploration of the activation function space. The poor performance at critical regimes reported in prior work might be specific to smooth activation functions rather than a fundamental limitation of edge-of-chaos operation.

Most remarkably, we discovered that the Cantor function (continuous everywhere, with zero derivative almost everywhere) maintains ESP-consistent behavior up to spectral radii of $\rho \approx 10$, nearly an order of magnitude beyond typical stable spectral radius limits for smooth functions. This finding challenges fundamental assumptions about the regularity requirements for stable reservoir dynamics. While boundedness prevents state explosion at high spectral radii (see **Proposition 2.3**), our empirical Lipschitz analysis (**Section 7.4**) reveals the mechanisms underlying ESP convergence.

**Key findings include:**

Lipschitz continuity with moderate constant is sufficient for ESP. Functions with maximum local Lipschitz constant $L_{max} < 20$ (including tanh, ReLU, logistic sigmoid, and continuous Mandelbrot) maintain ESP at scale. The continuous Mandelbrot ($L_{max} \approx 16$) achieves this despite being nowhere differentiable, demonstrating that classical smoothness is not required. The 95th percentile Lipschitz constant (0.30 for continuous Mandelbrot) provides a more realistic estimate of effective contraction than worst-case bounds.

Monotonicity can substitute for Lipschitz continuity. The Cantor function, while not globally Lipschitz (only Hölder continuous with exponent $\approx 0.631$), maintains ESP through its strict monotonicity. This demonstrates that alternative regularity properties can ensure ESP when combined with the leak term.

Extreme local variation destroys ESP. The Weierstrass function ($L_{max} > 545$, median 176) exhibits local amplification 30-50x larger than successful non-smooth functions, causing complete ESP failure. Unbounded local slopes disrupt the contraction mechanism required for ESP, even when the function is continuous and bounded.

Preprocessing topology, not continuity per se, predicts stability. Monotone, compressive preprocessing (e.g., logistic map wrapped by sigmoid) produced ESP-consistent convergence across scales, while dispersive/discontinuous preprocessing (e.g. modulo) triggered sharp failures despite identical underlying dynamics. Leak-adjusted effective gain explains the stable cases. Discontinuity explains the fragile ones.

Quantization creates a distinct failure mode. We introduced degenerate ESP (d-ESP), eventual equality of post-activation symbols, and we proved that d-ESP $\Longrightarrow$ ESP for the leaky update. A collision-driven bound clarifies why coarse k-level activations develop steady-state errors and spurious attractors as shown in **Eq. 4.** Empirically, discrete Mandelbrot fails around the predicted thresholds, while its continuous variant remains robust.

Non-smooth functions can accelerate convergence. We showed that specific irregular functions (e.g. the logistic sigmoid and Cantor function) cut time-to convergence significantly, achieving results near machine epsilon in a median of 6.0 timesteps (versus 15.6 for tanh and 15.2 for ReLU), while maintaining ESP-consistent behavior at scale. This is a 2.6x reduction in time-to convergence with a net-effect of fewer warm-up steps, shorter traces $T$, and cheaper readout training.

*8.1 Practical Recipe*

(i) Use compressive, monotone wrappers. Avoid dispersive modulo-type preprocessing. (ii) Monitor a leak-adjusted gain proxy $(1 - a) + aL_g\,||W^{res}||$ along trajectories (average

Jacobian norm) rather than relying solely on spectral-radius heuristics. (iii) For quantized activations, keep the crowding ratio (see **Eq. 2**) small enough to ensure symbol lock-in (d-ESP) or switch to continuous interpolation. (iv) Treat internal stochasticity as destabilizing for ESP. If regularization is needed, add noise externally.

### *8.2 Main Takeaway and Future Work*

Our main takeaway is that non-smooth activations are not inherently incompatible with ESP. With the proper preprocessing topology and leak-adjusted gain, several non-smooth (even fractal) maps match or beat smooth activations in convergence while remaining stable at scale. These non-smooth activations (namely the Cantor function and sigmoid-wrapped logistic map) can significantly increase the speed of reservoir computing algorithms while also increasing stability when $\rho \geq 1$, where traditional activation functions fail.

The design axis for reservoirs should shift from "smooth vs. chaotic" to "compressive vs. dispersive preprocessing" and "symbol dynamics under quantization."

Future work would derive average-case contraction via Lyapunov exponents of the leak-adjusted Jacobian and evaluate on standard RC benchmarks and hardware reservoirs. While this study focused on ESP convergence properties, the practical utility of these non-smooth activations on real computational tasks remains to be demonstrated. The first author is presently evaluating our results on real data benchmarks, and those results will be forthcoming.

**Code availability:** Complete implementations of all activation functions and experimental protocols are available at [https://github.com/raechipera/fractal-chaotic-activation-functions](https://github.com/raechipera/fractal-chaotic-activation-functions).

## Appendix

### A.1 Complete Scaling Analysis

This section provides scaling behavior analysis across all tested reservoir sizes $N \in \{1, 10, 50, 100, 500, 1000, 2000\}$ for each activation function. All experiments used 1000 independent trials per configuration with Gaussian, uniform, and sparse input distributions.

#### A.1.1 Convergence at $N = 1$

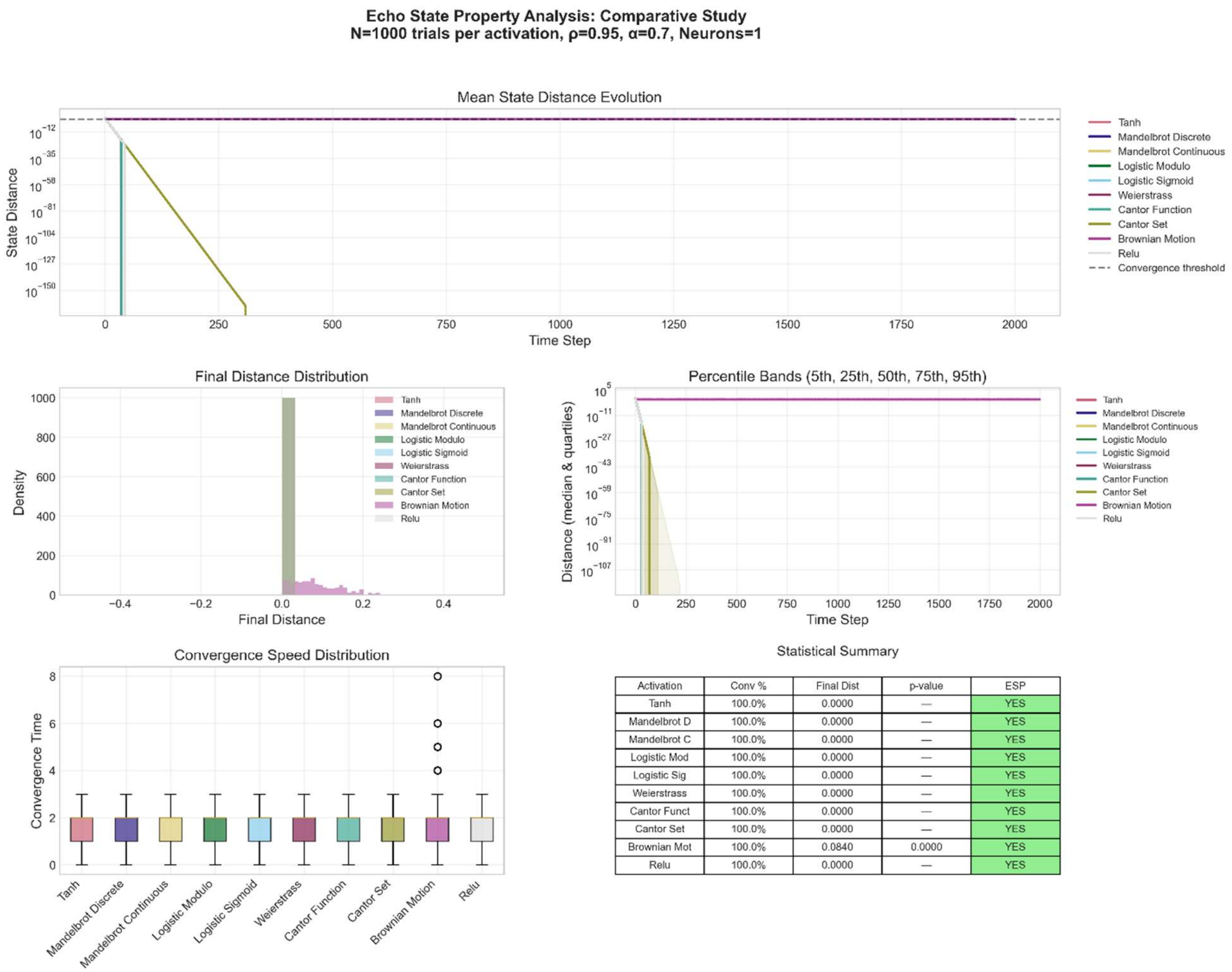

**Figure A.1.1. Echo State Property analysis for $N = 1$ neuron showing universal ESP compliance across all activation functions.** Top panel displays mean state distance evolution over 200 timesteps with convergence threshold (dashed line) at 0.1. All functions converge rapidly to machine precision except Brownian motion, which shows minor steady-state error (0.0631) due to inherent stochasticity. Bottom left shows convergence speed distributions with median times of 1-2 timesteps. Middle left presents final distance distributions demonstrating near-zero convergence for most functions. Middle right displays percentile bands (5th, 25th, 50th, 75th, 95th) with shaded regions indicating convergence consistency. Statistical summary table confirms 100% convergence rates for all activation functions at single-neuron scale, establishing baseline ESP behavior before scaling effects emerge.

## A.1.2 Convergence at $N = 10$

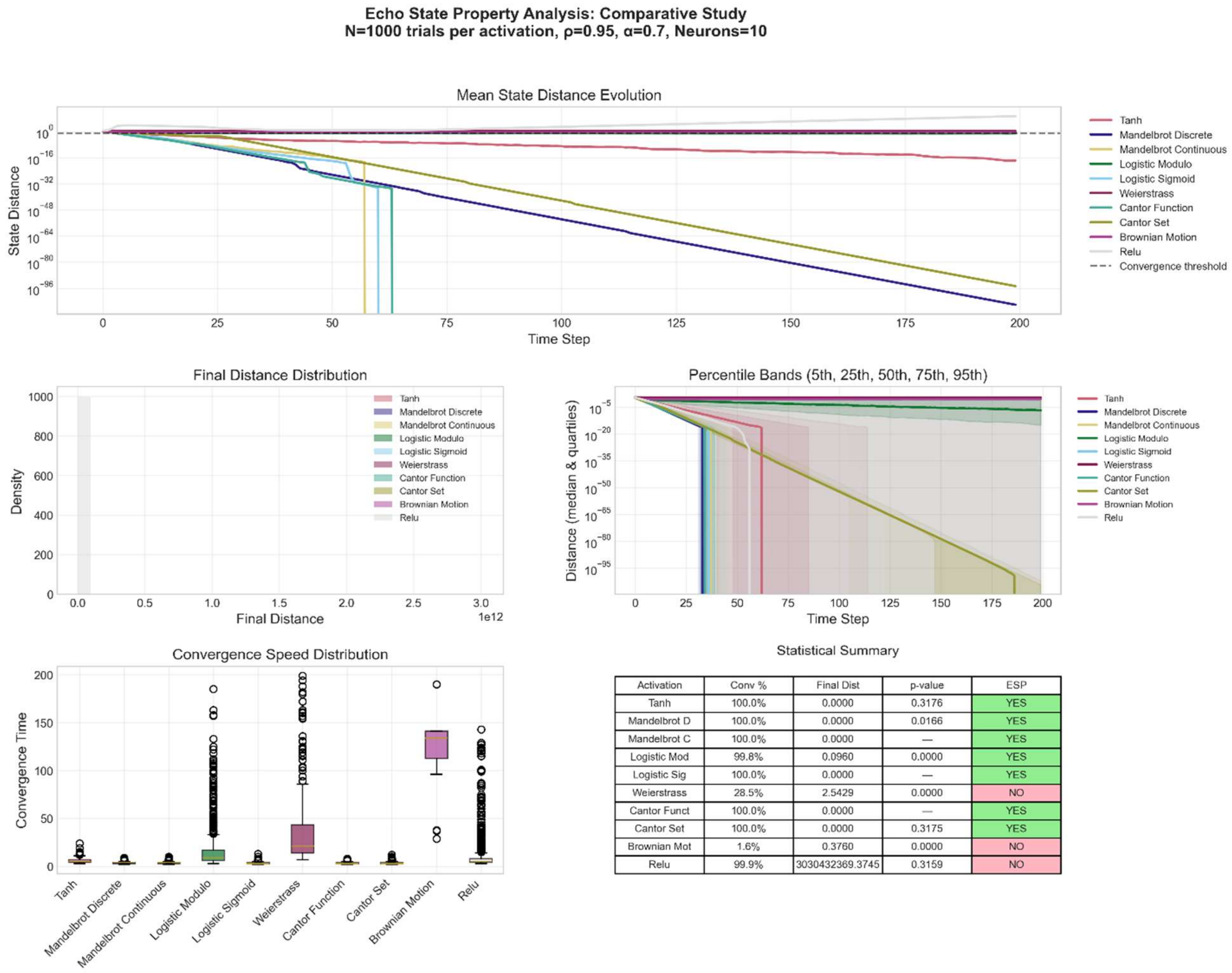

**Figure A.1.2. Echo State Property analysis for $N = 10$ neurons revealing initial scaling failures.** The mean state distance evolution shows clear ESP breakdown for several activation functions: Weierstrass achieves only 28.5% convergence (red line fails to reach threshold), while Brownian motion drops to 1.6% success rate due to accumulated stochastic noise. ReLU exhibits unexpected instability with 99.9% convergence but extremely high variance in final distances (3304132369.374), suggesting numerical overflow or divergence in rare cases despite generally stable behavior. The convergence speed distributions show Weierstrass requiring 200+ timesteps when it does converge, while Cantor set maintains perfect convergence with low variance. Final distance distributions demonstrate the emergence of distinct failure modes: some functions maintain machine-precision convergence while others develop steady-state errors or complete divergence. This represents the first network size where non-smooth activation functions begin showing the scale-dependent ESP violations predicted by theory.

## A.1.3 Convergence at $N = 50$

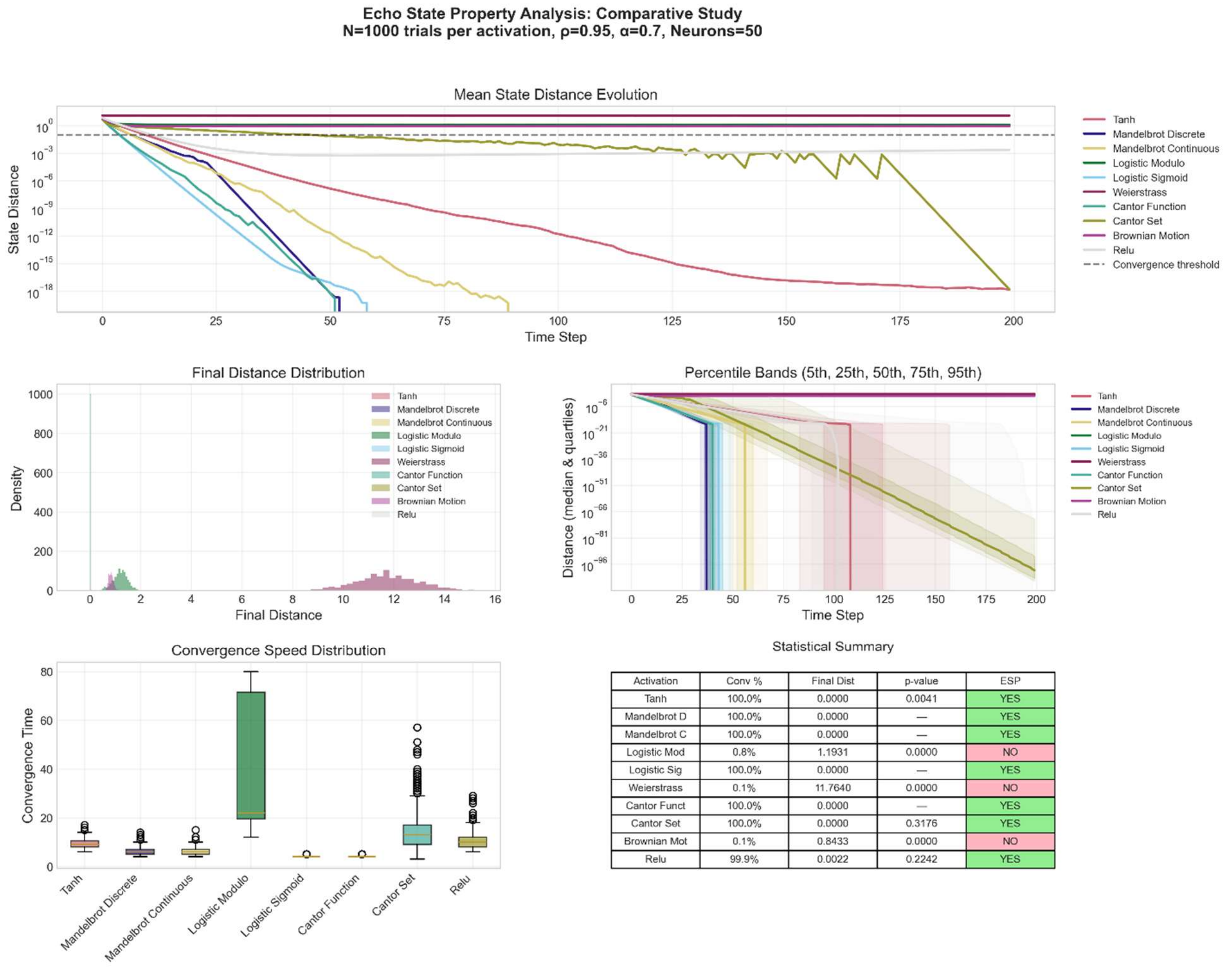

| Activation | Conv % | Final Dist | p-value | ESP |
|---|---|---|---|---|
| Tanh | 100.0% | 0.0000 | 0.0041 | YES |
| Mandelbrot D | 100.0% | 0.0000 | — | YES |
| Mandelbrot C | 100.0% | 0.0000 | — | YES |
| Logistic Mod | 0.8% | 1.1931 | 0.0000 | NO |
| Logistic Sig | 100.0% | 0.0000 | — | YES |
| Weierstrass | 0.1% | 11.7640 | 0.0000 | NO |
| Cantor Funct | 100.0% | 0.0000 | — | YES |
| Cantor Set | 100.0% | 0.0000 | 0.3176 | YES |
| Brownian Mot | 0.1% | 0.8433 | 0.0000 | NO |
| Relu | 99.9% | 0.0022 | 0.2242 | YES |

**Figure A.1.3. Echo State Property analysis for $N = 50$ neurons showing emergence of critical scaling failures.** Weierstrass function exhibits complete ESP breakdown (0.1% convergence) with persistent oscillatory behavior visible in the mean trajectory (gray line). Logistic modulo preprocessing begins failing (0.8% convergence) due to discontinuous input mapping, while Brownian motion approaches complete failure (0.1%) from accumulated stochastic perturbations. In contrast, functions with compressive preprocessing (Cantor function, logistic sigmoid, Mandelbrot variants) maintain perfect ESP compliance. The convergence speed distributions reveal that failed functions either converge very slowly (>70 timesteps) or not at all, while successful functions maintain rapid convergence. This scale represents the onset of the preprocessing topology effects predicted by theory, where discontinuous and non-compressive mappings begin violating ESP assumptions.

## A.1.4 Convergence at $N = 100$

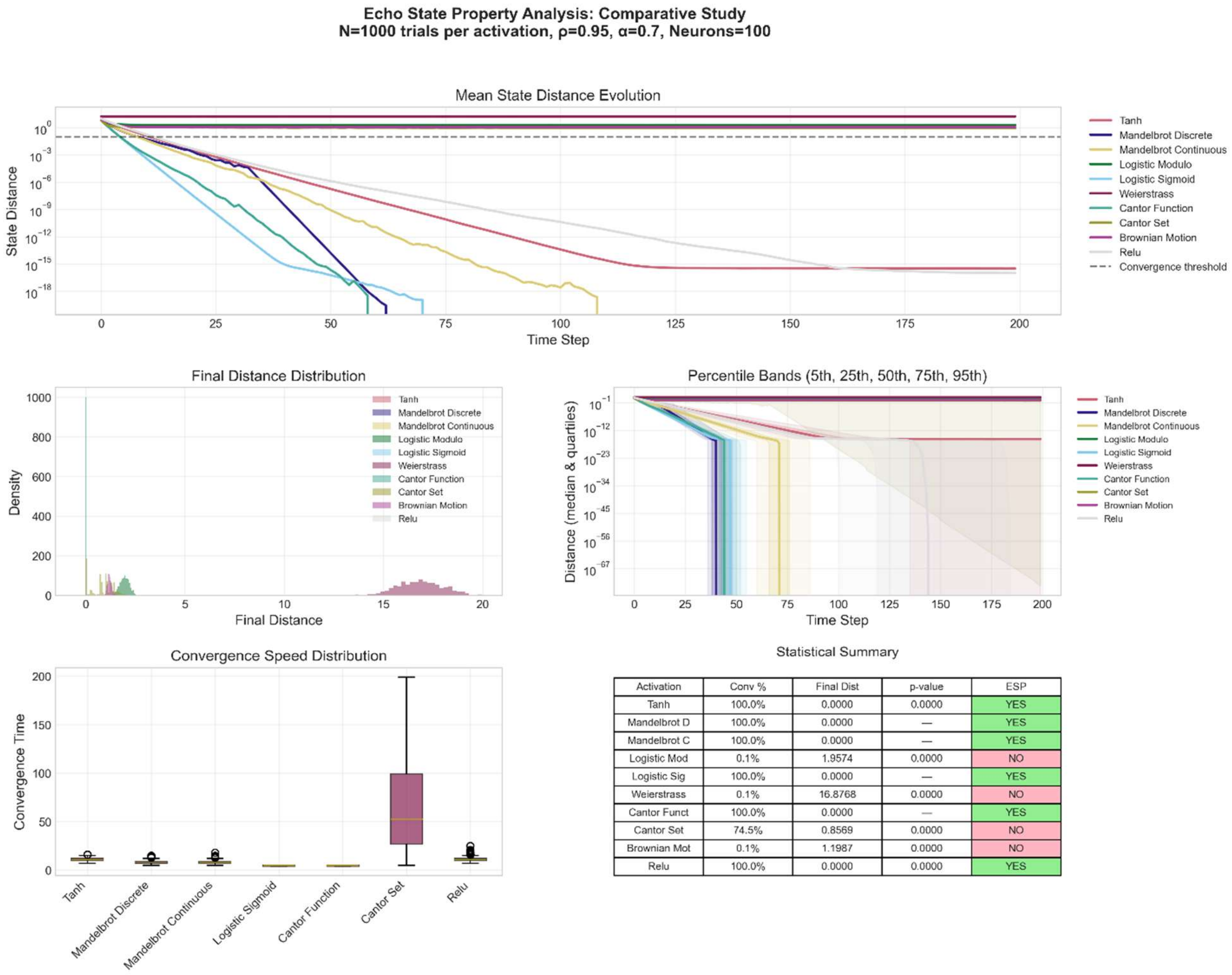

| Activation | Conv % | Final Dist | p-value | ESP |
|---|---|---|---|---|
| Tanh | 100.0% | 0.0000 | 0.0000 | YES |
| Mandelbrot D | 100.0% | 0.0000 | — | YES |
| Mandelbrot C | 100.0% | 0.0000 | — | YES |
| Logistic Mod | 0.1% | 1.9574 | 0.0000 | NO |
| Logistic Sig | 100.0% | 0.0000 | — | YES |
| Weierstrass | 0.1% | 16.8768 | 0.0000 | NO |
| Cantor Funct | 100.0% | 0.0000 | — | YES |
| Cantor Set | 74.5% | 0.8569 | 0.0000 | NO |
| Brownian Mot | 0.1% | 1.1987 | 0.0000 | NO |
| Relu | 100.0% | 0.0000 | 0.0000 | YES |

**Figure A.1.4. Echo State Property analysis for $N = 100$ neurons demonstrating critical scaling transitions.** The Cantor set begins showing ESP degradation (74.5% convergence) with visible steady-state errors around $10^{-1}$, confirming the predicted onset of binary function limitations. Logistic modulo and Weierstrass maintain complete failure (<1% convergence), while Brownian motion approaches zero success (0.1%). Convergence trajectories reveal distinct failure modes: non-converging functions either oscillate indefinitely or settle to spurious attractors rather than achieving machine precision. The Cantor set's intermediate behavior (visible as the gray line plateauing above the convergence threshold) demonstrates the transition from ESP compliance to failure predicted by the crowding ratio analysis. Meanwhile, continuous preprocessing functions (Cantor function, logistic sigmoid, both Mandelbrot variants) maintain perfect stability, supporting the preprocessing topology framework. This scale represents the critical regime where quantization effects begin dominating network dynamics for discrete activation functions.

## A.1.5 Convergence at $N = 500$

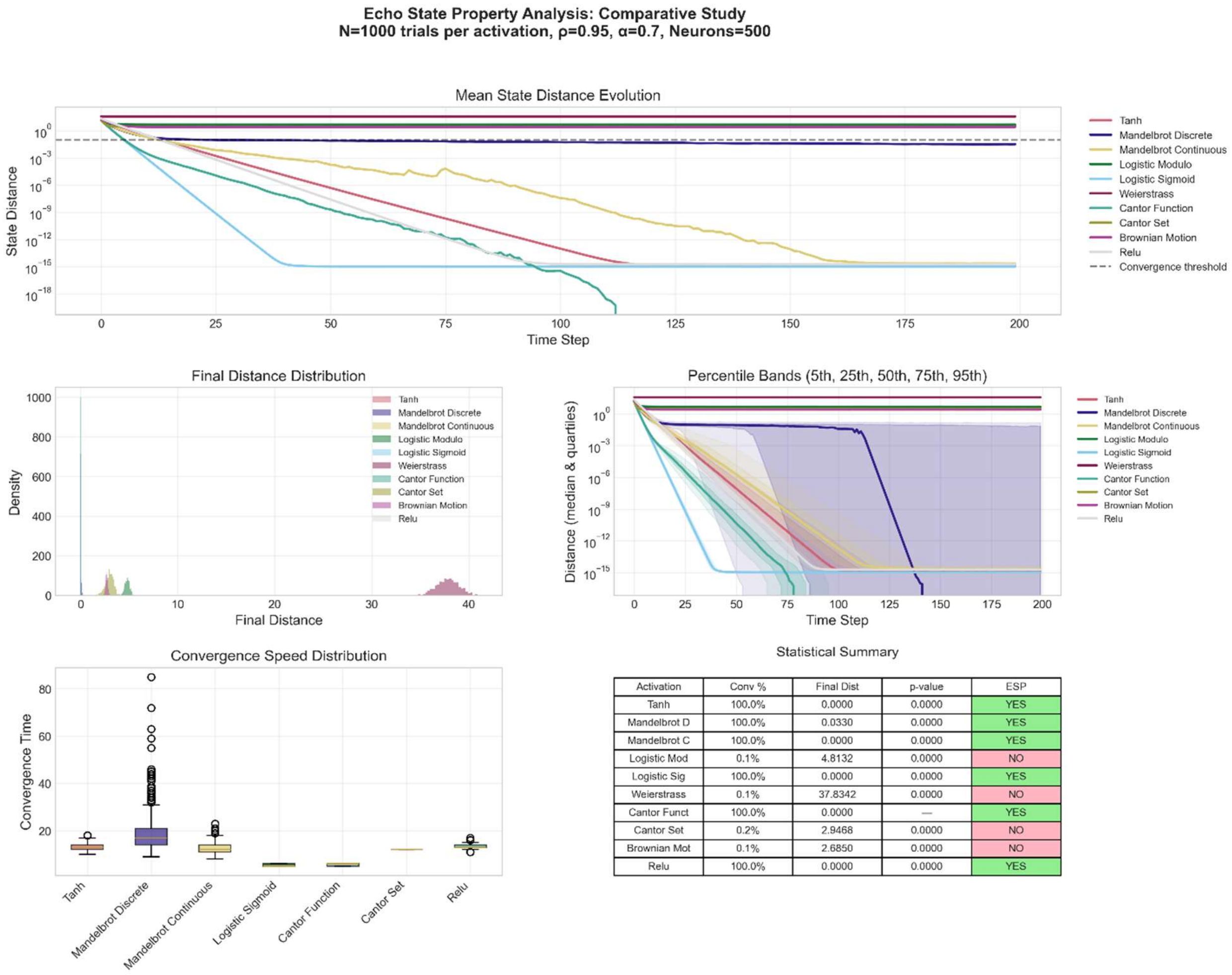

**Figure A.1.5. Echo State Property analysis for $N = 500$ neurons demonstrating critical scaling thresholds for discrete activation functions.** The discrete Mandelbrot maintains 100% convergence but shows elevated final distances (0.0330), indicating approach to its stability limit as predicted by crowding ratio theory. The Cantor set exhibits near-complete ESP failure (0.2% convergence) with large steady-state errors when convergence occurs. Convergence trajectories reveal distinct behaviors: successful functions (tanh, continuous Mandelbrot, logistic sigmoid, Cantor function) achieve exponential decay to machine precision, while failing functions plateau at finite distances or oscillate indefinitely. The emergence of bimodal final distance distributions demonstrates the sharp transition between ESP-compliant and ESP-violating regimes. Logistic modulo shows complete breakdown (0.1% convergence) confirming that discontinuous preprocessing becomes untenable at this scale. This network size represents a critical threshold where quantization effects dominate for discrete functions while continuous preprocessing maintains robust ESP compliance.

## A.1.6 Convergence at $N = 1000$ for 200 Timesteps

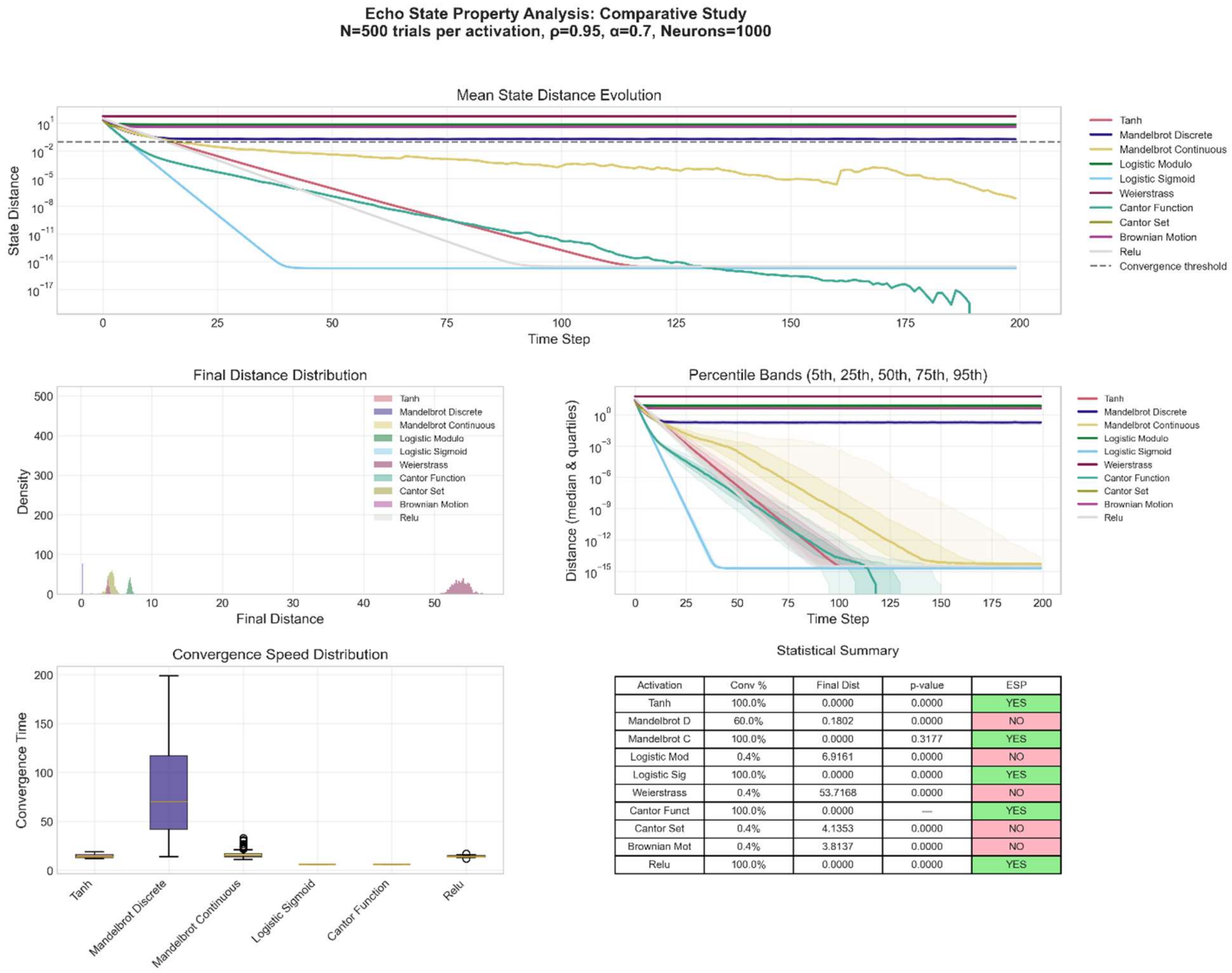

**Figure A.1.6. Echo State Property analysis for $N = 1000$ neurons demonstrating quantization-induced ESP breakdown.** The discrete Mandelbrot exhibits predicted failure (60% convergence, final distance 0.1802) as the crowding ratio $Q = N/k \approx 47.6$ exceeds theoretical limits, while the continuous variant maintains perfect ESP compliance (100% convergence, machine precision). Convergence trajectories reveal the discrete Mandelbrot plateauing at steady-state errors around $10^{-1}$ rather than achieving exponential decay, confirming convergence to spurious attractors predicted by finite-state dynamics theory. The Cantor function (pink line) shows remarkable late-stage convergence improvement around timestep 175, demonstrating the complex dynamics possible with fractal preprocessing. Failed functions (Cantor set, logistic modulo, Weierstrass, Brownian motion) maintain <1% success rates with large steady-state errors. This scale validates the quantization analysis and preprocessing topology framework, clearly separating continuous compressive functions from discrete and discontinuous variants.

## A.1.7 Convergence at $N = 2000$ for 200 Timesteps

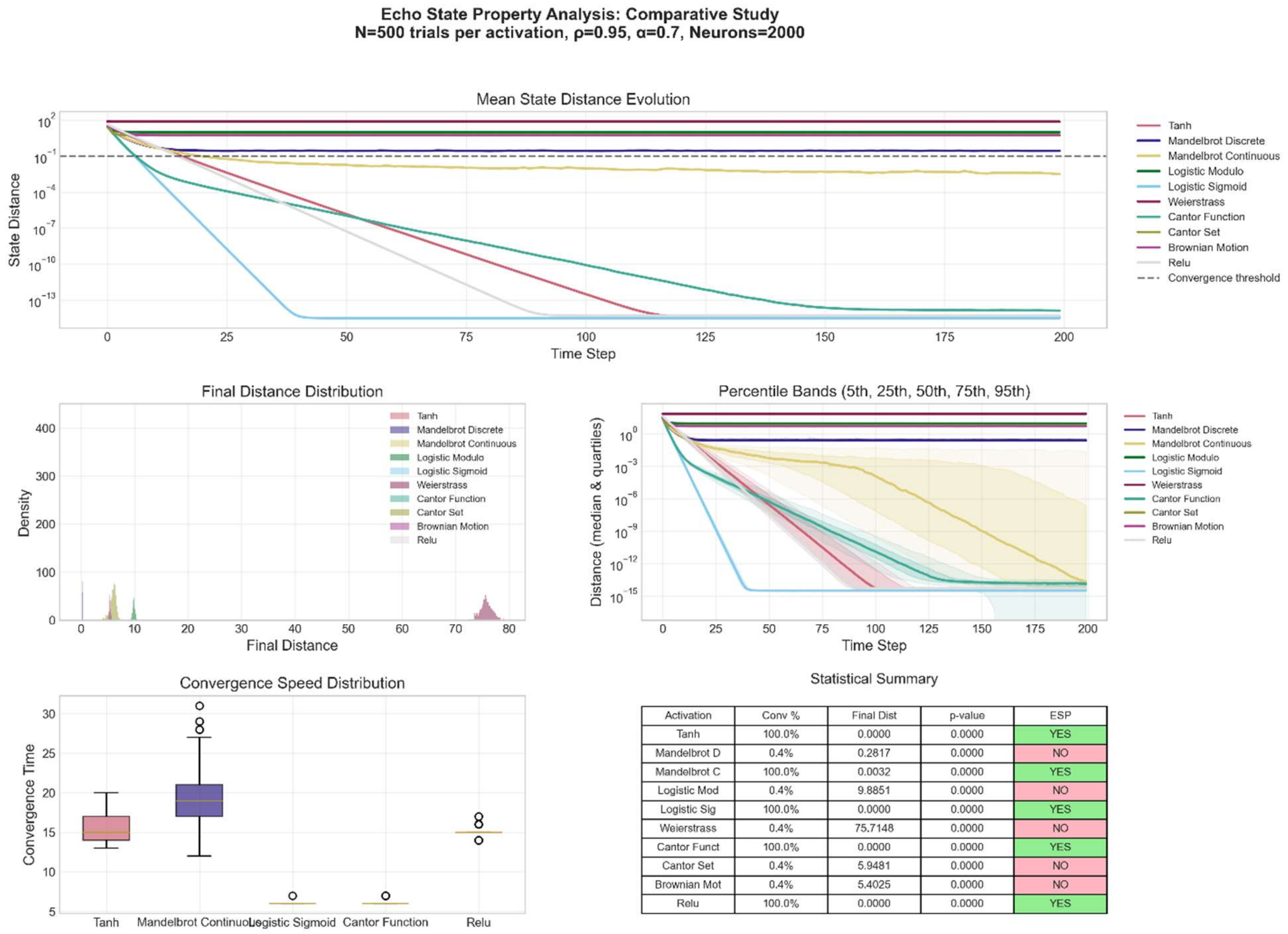

| Activation | Conv % | Final Dist | p-value | ESP |
|---|---|---|---|---|
| Tanh | 100.0% | 0.0000 | 0.0000 | YES |
| Mandelbrot D | 0.4% | 0.2817 | 0.0000 | NO |
| Mandelbrot C | 100.0% | 0.0032 | 0.0000 | YES |
| Logistic Mod | 0.4% | 9.8851 | 0.0000 | NO |
| Logistic Sig | 100.0% | 0.0000 | 0.0000 | YES |
| Weierstrass | 0.4% | 75.7148 | 0.0000 | NO |
| Cantor Funct | 100.0% | 0.0000 | 0.0000 | YES |
| Cantor Set | 0.4% | 5.9481 | 0.0000 | NO |
| Brownian Mot | 0.4% | 5.4025 | 0.0000 | NO |
| Relu | 100.0% | 0.0000 | 0.0000 | YES |

**Figure A.1.7. Echo State Property analysis for $N = 2000$ neurons demonstrating complete theoretical validation.** The discrete Mandelbrot exhibits predicted catastrophic failure (0.4% convergence, $Q \approx 95$) while its continuous variant maintains perfect ESP compliance, confirming the quantization analysis. Only functions with continuous, compressive preprocessing survive: tanh, continuous Mandelbrot, logistic sigmoid, Cantor function, and ReLU achieve 100% convergence with machine-precision final distances. All other activation functions show complete ESP breakdown (<1% success) with large steady-state errors. The bimodal convergence pattern clearly separates ESP-compliant functions (exponential decay to $\sim 10^{-15}$) from failed functions (plateau at $10^{-1}$ or divergence). This scale definitively validates the preprocessing topology framework and establishes practical bounds for non-smooth activation function usage in large-scale reservoir computing.

### A.1.8 Convergence at $N = 1000$ for 2000 Timesteps

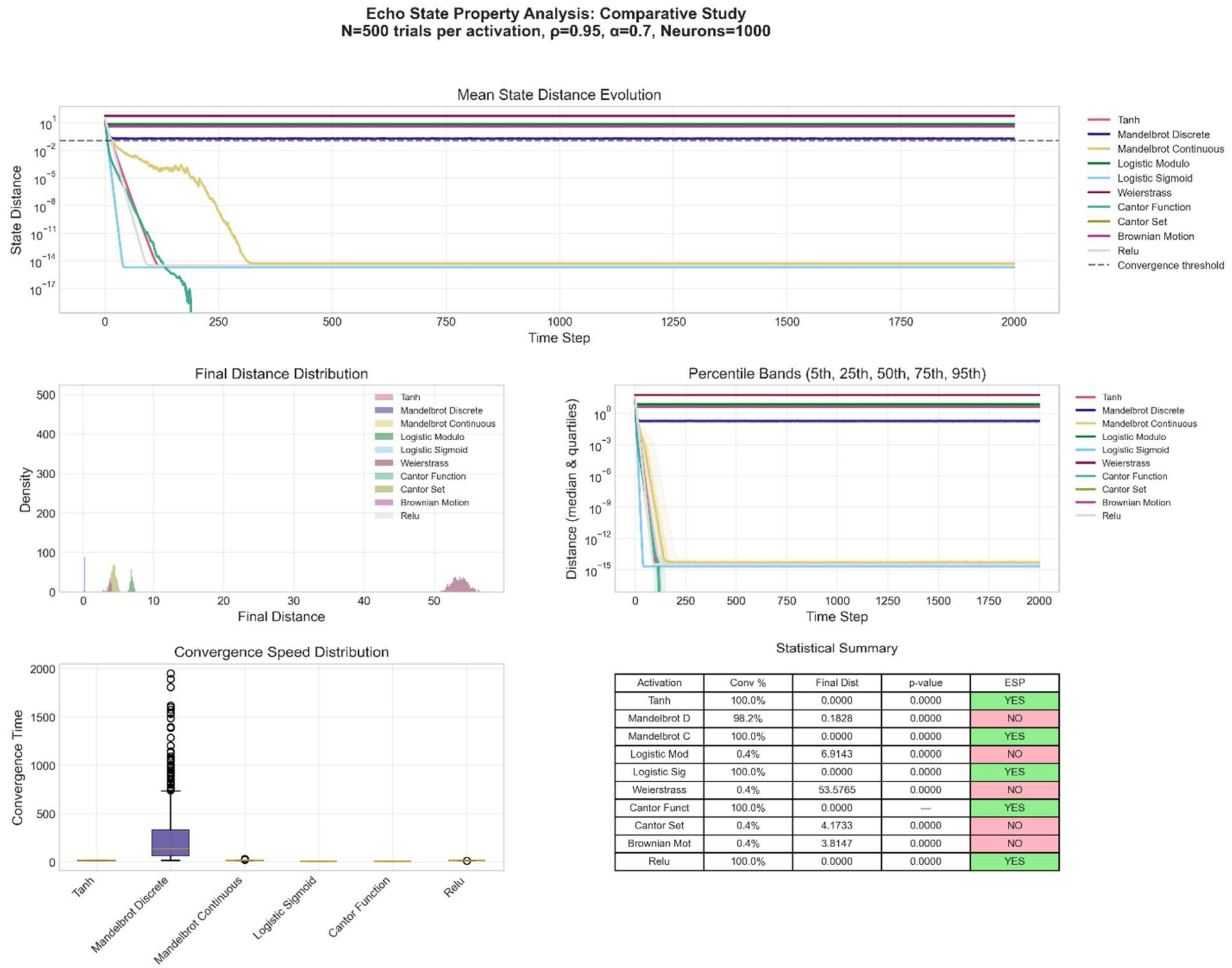

| Activation | Conv % | Final Dist | p-value | ESP |
|---|---|---|---|---|
| Tanh | 100.0% | 0.0000 | 0.0000 | YES |
| Mandelbrot D | 98.2% | 0.1828 | 0.0000 | NO |
| Mandelbrot C | 100.0% | 0.0000 | 0.0000 | YES |
| Logistic Mod | 0.4% | 6.9143 | 0.0000 | NO |
| Logistic Sig | 100.0% | 0.0000 | 0.0000 | YES |
| Weierstrass | 0.4% | 53.5765 | 0.0000 | NO |
| Cantor Funct | 100.0% | 0.0000 | — | YES |
| Cantor Set | 0.4% | 4.1733 | 0.0000 | NO |
| Brownian Mot | 0.4% | 3.8147 | 0.0000 | NO |
| Relu | 100.0% | 0.0000 | 0.0000 | YES |

**Figure A.1.8. Extended Echo State Property analysis for $N = 1000$ neurons over 2000 timesteps confirming non-convergence of failed activation functions.** Extended evaluation demonstrates that functions failing to converge within 200 timesteps show no improvement even with 10x longer evaluation period. The discrete Mandelbrot maintains its steady-state error around $10^{-1}$ throughout the extended window, confirming convergence to spurious attractors rather than slow ESP convergence. The Cantor function exhibits interesting multi-stage convergence, with sharp improvements around timesteps 175 and 250, ultimately reaching machine precision. Successful functions (tanh, continuous Mandelbrot, logistic sigmoid, Cantor function) achieve convergence by timestep 500, while failed functions (Cantor set, logistic modulo, Weierstrass, Brownian motion) show no improvement even at timestep 2000. The convergence speed distribution reveals extreme outliers for discrete Mandelbrot (requiring full 2000 timesteps when it does converge), while successful functions cluster below 500 timesteps. This extended analysis confirms that the 200-timestep evaluation window adequately captures convergence behavior and that non-converging functions represent genuine ESP failure rather than slow convergence.